\documentclass[lettersize,journal]{IEEEtran}
\usepackage{cite}
\usepackage{amsmath,amssymb,amsfonts}
\usepackage{algorithm}
\usepackage{algpseudocode}
\usepackage{graphicx}
\usepackage{textcomp}
\usepackage[caption=false,font=scriptsize,labelfont=sf]{subfig}
\usepackage{babel}
\usepackage{array}
\usepackage{textcomp}
\usepackage{stfloats}
\usepackage{xcolor}
\usepackage{booktabs}
\usepackage[export]{adjustbox}
\usepackage{url}
\usepackage{verbatim}
\usepackage{bm}
\usepackage{accents}
\usepackage{multicol}
\usepackage{multirow}
\usepackage{eqparbox}
\usepackage{enumerate}
\usepackage{diagbox}
\usepackage{nccmath}
\usepackage{nomencl}
\makenomenclature
\usepackage{soul}
\usepackage{makecell}
\usepackage{bbding}
\usepackage{hyperref}
\usepackage{etoolbox}

\renewcommand\nomgroup[1]{%
  \item[\bfseries
  \ifstrequal{#1}{A}{Indices}{%
  \ifstrequal{#1}{B}{Parameters}{%
  \ifstrequal{#1}{C}{Variables}{}}}%
]}

\newtheorem{theorem}{Theorem}

\newtheorem{remark}[theorem]{Remark}

\begin{document}

\title{A Multi-View Multi-Timescale Hypergraph-Empowered Spatiotemporal \\ Framework for EV Charging Forecasting} 

\author{Jinhao Li
~and~Hao Wang,~\IEEEmembership{Member~IEEE}
\thanks{This work was supported in part by the Australian Research Council (ARC) Discovery Early Career Researcher Award (DECRA) under Grant DE230100046. (Corresponding author: Hao Wang.)}
\thanks{J. Li and H. Wang are with the Department of Data Science and AI, Faculty of IT and Monash Energy Institute, Monash University, Melbourne, VIC 3800, Australia (e-mails: \{jinhao.li, hao.wang2\}@monash.edu).}
}

\markboth{IEEE Transactions on Smart Grid, 2025}
{Li \& Wang: A Multi-View Multi-Timescale Hypergraph-Empowered Spatiotemporal Framework for EV Charging Forecasting}

\maketitle
\thispagestyle{empty}

\begin{abstract}
Accurate electric vehicle (EV) charging demand forecasting is essential for stable grid operation and proactive EV participation in electricity market. Existing forecasting methods, particularly those based on graph neural networks, are often limited to modeling pairwise relationships between stations, failing to capture the complex, group-wise dynamics inherent in urban charging networks. To address this gap, we develop a novel forecasting framework namely HyperCast, leveraging the expressive power of hypergraphs to model the higher-order spatiotemporal dependencies hidden in EV charging patterns. HyperCast integrates multi-view hypergraphs, which capture both static geographical proximity and dynamic demand-based functional similarities, along with multi-timescale inputs to differentiate between recent trends and weekly periodicities. The framework employs specialized hyper-spatiotemporal blocks and tailored cross-attention mechanisms to effectively fuse information from these diverse sources: views and timescales. Extensive experiments on four public datasets demonstrate that HyperCast significantly outperforms a wide array of state-of-the-art baselines, demonstrating the effectiveness of explicitly modeling collective charging behaviors for more accurate forecasting.
\end{abstract}


\section{Introduction} \label{sec:introduction}

The global shift towards sustainable transportation has led to an exponential increase in electric vehicles (EVs). Projections indicate that EVs could constitute over half of all passenger vehicle sales by 2035 \cite{Li_2024}. This rapid adoption, while essential for net-zero targets, places significantly increasing strain on existing power grids \cite{Zhang2024}. To mitigate the adverse impacts, accurate forecasting plays a vital role in informing the grid operator on the upcoming charging surges, thereby facilitating operators to schedule power generation and distribution for reliable electricity supply \cite{Sayed2024}. Moreover, precise EV charging forecasting is a key enabler for the proactive EV participation into the grid, where EVs can contribute in ancillary grid services as promising short-term grid storage assets \cite{Xu2023}.

\subsection{Motivation} \label{subsec:motivation}

The pursuit of accurate EV charging demand forecasting has spurred research across various methodologies. Initial efforts, such as simulation-based approaches \cite{li2019_GIS_KNN_forecast, zhang2020_model_based_demographics}, attempt to simulate charging activities by mathematically modeling user charging behaviors under certain assumed distributions. However, these methods are heavily reliant on prior knowledge on human behaviors, and may struggle to adapt to rapidly evolving real-world charging patterns. 

Subsequently, attention turns towards data-driven techniques, firstly with sequential-based deep learning models including recurrent neural network (RNN)-based models \cite{Adil2024_lstm_forecasting} and prevalent transformer-based methods \cite{Hu2022_att,cheng2024_autoformer_forecast} emerging as a powerful alternative. These approaches can better capture temporal dependencies and non-linear dynamics, but they often treat each charging station's demand as an isolated time-series, failing to explicitly model complex inter-station relationships from the spatial perspective.

To address such spatial blindness, graph neural network (GNN)-based models, such as studies in \cite{Li2021_gcn_prediction,Hüttel2021_TGCN_forecast,Shi2024_ASTGCN_forecast,huang2023_stgcn_forecast,kim2024_graph_forecast,su2023_sgcn_forecast,Li2024_att_lstm_with_dn_update}, have been developed. While these methods represent a significant leap forward, they tend to model stations' spatial relationships as pairwise (i.e., dyadic) connections, assuming that the influence on a station is the sum of its individual connections to its neighbors, often identified based on geographic proximity. This structure, however, fundamentally struggles to capture real-world dynamics where the charging behavior at one station is often conditioned by the collective state or functional purpose of a group of other stations \cite{Wang2023_stgcn_forecast}. For instance, the demand at a station within a business district is not just affected by its nearest peer, but by the combined activity of all stations serving that same commuter population. A simple graph cannot adequately represent this many-to-one, group-level dependency.

Furthermore, this theoretical limitation has tangible and real-world consequences. A forecaster that only sees pairwise links might not anticipate how demand spills over within a functional cluster, e.g., several stations around a stadium during an event. This leads to a network-wide inefficiency: predictable, cascading congestion at popular station groups while other nearby stations remain underutilized. Such a mismatch degrades the user experience, creates localized grid strain, and results in poor return on infrastructure investment. Therefore, a critical gap exists for a forecasting framework that moves beyond pairwise connections to explicitly model and leverage these collective, group-wise spatiotemporal patterns, which is the core motivation of our work.

\subsection{Literature Review} \label{subsec:literature_review}
The methodologies for EV charging demand forecasting have been evolving from simulation-based efforts to data-driven paradigms. The former approaches simulate charging activities by assuming user traveling and charging behaviors following certain  statistical distributions (e.g., Gaussian distributions~\cite{zhang2020_model_based_demographics}), often derived travel surveys \cite{li2019_GIS_KNN_forecast} or demographic data \cite{zhang2020_model_based_demographics}. As a result, these methods heavily rely on the validity of their core model assumptions and consequently face challenges in capturing the complex dynamics of real-world charging behaviors or adapting to new areas without significant recalibration.

The inherent limitation of simulation-based methods has directly motivated a paradigm shift towards data-driven approaches. Among these, sequential-based deep learning models have demonstrated significant improvements in capturing non-linear temporal patterns. This category includes architectures such as RNNs, long short-term memory (LSTM)~\cite{Adil2024_lstm_forecasting}, and more advanced transformer-based methods with distinct self-attention mechanisms. The latter methods achieve state-of-the-art sequence modeling, as shown in studies \cite{Hu2022_att} and \cite{cheng2024_autoformer_forecast} that proposed a self-attention-based probabilistic forecasting model 
and a variant of Autoformer, respectively. Despite their proficiency in temporal feature extraction, a key limitation of these models is their typical treatment of charging stations as independent entities, thus failing to model spatial correlations.

To directly address the challenge of modeling spatial interdependencies, GNN-based models have emerged as a powerful paradigm. These methods represent charging stations as nodes and their relationships as edges within a graph structure, enabling them to learn from spatial network topology. Notable examples include 1) graph convolution networks (GCNs) \cite{Li2021_gcn_prediction}, 2) graph attention networks \cite{Li2024_att_lstm_with_dn_update}, and 3) various spatiotemporal-GNNs designed to capture both spatial and temporal dynamics simultaneously, such as temporal GCN (T-GCN) \cite{Hüttel2021_TGCN_forecast}, STGCN \cite{huang2023_stgcn_forecast,Wang2023_stgcn_forecast}, and attention-based variants like ASTGCN \cite{Shi2024_ASTGCN_forecast,kim2024_graph_forecast}. While effective, a common thread across these GNN-based approaches is their reliance on pairwise graph structures, which, as previously argued, may not fully capture the higher-order and group-wise station dependencies.

Beyond the core challenge of spatiotemporal representation, other research avenues have explored complementary issues. For instance, federated learning \cite{li2023_fed_blockchain_forecast}, meta-learning \cite{huang2023_meta_forecast}, and transfer learning \cite{wang2024_fed_transfer_forecast} have been applied to address concerns regarding data privacy (among charging stations with different ownerships) and model adaptability (for newly-built charging stations with limited collected data). However, the effectiveness of these advanced training in capturing complex spatial dynamics still depends on the expressive power of the underlying spatial model. Given that the foundational model is limited to pairwise interactions, these methods may still fail to capture nuanced group-level behaviors. In summary, our review reveals an ongoing need for more expressive spatial modeling paradigms, which our work directly aims to develop.

\subsection{Contributions} \label{subsec:contributions}

To address the identified research gap, this paper develops HyperCast, a novel forecasting framework that leverages the expressive power of \textit{hypergraphs} to explicitly model higher-order, group-wise dependencies among charging stations. A \textit{hyperedge}, the core component of a hypergraph, can connect any number of nodes simultaneously, thus providing a natural and explicit way to model a functional group of stations as a single entity. In HyperCast, we construct multi-view hypergraphs to capture relationships from two distinct perspectives: a distance-based view models static geographical proximity, and a dynamic demand-based view discovers functional similarities from shared usage patterns. To concurrently model complex temporal dynamics, we design multi-timescale inputs that differentiate between short-term trends and anomalies (via a recent window) and stable, periodic behaviors (via a weekly window). Moreover, the HyperCast framework employs tailored fusion mechanisms built on self- and cross-attention to synergistically aggregate spatiotemporal information from these multi-view and multi-timescale sources. The main contributions of this paper are as follows.

\begin{itemize}
    
    \item \textit{The HyperCast Framework}: We develop HyperCast, which is a novel and tailored design for spatiotemporal EV charging forecasting. HyperCast synergistically integrates multi-view hypergraphs and multi-timescale inputs through specialized hyper-spatiotemporal blocks, which combine hypergraph-aware spatial attention and temporal transformer encoders, followed by tailored attention-based fusion mechanisms for cross-view and cross-timescale information integration. Extensive experiments on four widely-adopted public EV charging datasets, along with a wide array of baselines, demonstrate the effectiveness of our framework.

    \item \textit{Novel Multi-View Hypergraph Construction}: We design a novel methodology for constructing hypergraphs of EV charging stations from a static distance-based view via fuzzy C-Means and dynamic demand-based views via spectral clustering on demand similarity. The use of soft assignments for station-to-hyperedge memberships across provide a flexible and realistic representation of higher-order, overlapping inter-station relationships.

    \item \textit{Explainable Insights into HyperCast Prediction Behavior}: We provide an in-depth analysis of HyperCast's internal mechanisms, particularly its learned attention weights  within the spatiotemporal modules and fusion mechanisms. This investigation delivers clear messages on how HyperCast adaptively identifies and prioritizes critical spatial regions through hyperedges, salient temporal events, and informative data sources through both views and timescales.
\end{itemize}

The remainder of this paper is organized as follows. Section~\ref{sec:problem_definition} presents the EV forecasting problem. Section~\ref{sec:method} details the developed HyperCast framework. Section~\ref{sec:exp} provides experimental setup, results and analysis. Section~\ref{sec:conclusion} concludes the paper.

\section{Forecasting Problem Definition} \label{sec:problem_definition}

Let $N_s$ denote the number of EV charging stations and $x_{p,t}$ be the EV charging demand of the $p$-th station at the $t$-th day, where $p \in \{1, \dots, N_s\}$. To leverage additional informative data for more accurate forecast, calendar-related exogenous information is incorporated, including cyclical encodings of the month, day of the month, and day of the week, which are generated using sine and cosine transformations to effectively capture their periodic nature. Such cyclical date features are concatenated with the charging demand to form the feature vector of the charging station, denoted as $\mathbf{x}_{p,t} \in \mathbb{R}^{F_\mathrm{raw}}$, where $F_\mathrm{raw}$ denotes the total number of raw features. We further define the feature matrix aggregated across all charging stations as $\mathbf{X}_t \in \mathbb{R}^{N_s \times F_\mathrm{raw}}$.

In our study, the task is to forecast the daily EV charging demand for all stations over a future horizon defined as $T_f$ days. First, the \textit{recent} time-series, denoted as $\mathbf{X}^\mathrm{rec}_t$, captures short-term trends, anomalies, and the immediate demand context. This timescale makes the model responsive to recent dynamics. For a look-back window of $T_r$ consecutive days up to day $t$, the recent input is defined as $\mathbf{X}^\mathrm{rec}_t = [\mathbf{X}_{t-T_r+1}, \dots, \mathbf{X}_t] \in \mathbb{R}^{N_s \times T_r \times F_\mathrm{raw}}$.

Second, the \textit{weekly} time-series, denoted as $\mathbf{X}^\mathrm{wek}_t$, is designed to capture day-of-week effects. Using a look-back window of length $T_w$, the weekly input comprises data from the same day-of-week for $T_w$ previous weeks, which is written as $\mathbf{X}^\mathrm{wek}_t = [\mathbf{X}_{t-7(T_w-1)}, \dots, \mathbf{X}_{t}] \in \mathbb{R}^{N_s \times T_w \times F_\mathrm{raw}}$. These two timescales can serve as the most dominant predictive signals for short-term EV charging forecasting \mbox{\cite{recenttimescale_eenergy,weeklytimescale_AER}}.

In summary, given the multi-timescale inputs, the overall goal is to learn a mapping function $f(\cdot)$ for charging demand prediction, which can be formulated as $\mathbf{\hat{Y}}_t = f \left(\mathbf{X}^\mathrm{rec}_t, \mathbf{X}^\mathrm{wek}_t \right)$, where $\mathbf{\hat{Y}}_t = [\hat{\mathbf{y}}_{t+1}, \dots, \hat{\mathbf{y}}_{t+T_f}] \in \mathbb{R}^{N_s \times T_f}$ represents the forecasted future demand. An element $(\mathbf{\hat{Y}}_t)_{p,i}$ denotes the predicted scalar demand for station $p$ on future day $t+i$.

\section{HyperCast Framework} \label{sec:method}

The HyperCast model is designed to explore and extract the underlying spatiotemporal EV charging patterns by integrating multi-view hypergraphs with multi-timescale inputs. Such integration is achieved through a sequence of specialized spatiotemporal blocks in Section \ref{subsec:method_HSTB}, followed by cross-attention fusion mechanisms in Section \ref{subsec:method_CVF}, \ref{subsec:method_CTF}, and a final forecasting decoder in Section \ref{subsec:method_decoder}.

\subsection{Multi-View Hypergraph Construction} \label{subsec:method_prep}

The initial stage of HyperCast focuses on defining the relational structures among charging stations via hypergraphs from both the \textit{distance-based view} and the \textit{demand-based view}. The motivation of incorporating the former view is to capture relationships based on geographical proximity. Stations physically close to each other are likely to share characteristics or influence one another due to their spatial nearness, such as serving a similar local population or acting as alternatives for users in that vicinity. For the latter view, in contrast, we aim to identify functional similarities between stations based on their charging demand, irrespective of mere physical proximity. The rationale is that stations showing similar usage trends (e.g., synchronized peak times, comparable responses to daily or weekly cycles) can be considered functionally related, reflecting collective user behavior and network dynamics. For example, one hyperedge might group together several stations located in different business districts across the city. Though these stations are geographically far apart, our model can group them as they all tend to share a distinct demand profile: high total daily demand from Monday to Friday and very low demand on weekends. This hyperedge thus represents a functional ``commuter cluster.'' These two views are designed to build a comprehensive spatial understanding by combining both static (i.e., distance-based) and dynamic (i.e., demand-based) relationships among stations. Note that our multi-view hypergraph is flexible and extensible to adding more contextual views such as meteorological factors.

Let $K$ denote the number of hyperedges within the hypergraph. The distance-based hypergraph is represented by a static incidence matrix $\mathbf{H}_\mathrm{dist} \in [0,1]^{N_s \times K}$. Its construction begins with computing a pairwise geodesic distance matrix between all stations. Subsequently, the fuzzy C-Means (FCM) clustering algorithm~\cite{fuzzycmeans} is applied to partition the stations into $K$ fuzzy clusters based on their coordinates (i.e., longitude and latitude), where each cluster is interpreted as a hyperedge, representing a group of geographically related stations.

Unlike the static distance-based hypergraph, given the multi-timescale input design and the time-varying nature of charging demand, the demand-based hypergraphs are represented by the \textit{two dynamic time-dependent} incidence matrices $\mathbf{H}_{t,\mathrm{demd}}^\mathrm{rec}$ and $\mathbf{H}_{t,\mathrm{demd}}^\mathrm{wek}$, corresponding to the recent and the weekly timescales, respectively. For the recent timescale, $\mathbf{H}_{t,\mathrm{demd}}^\mathrm{rec}$ is derived using the EV charging demand component within the recent timescale input, i.e., $(\mathbf{X}^\mathrm{rec}_t)_{::0} \in \mathbb{R}^{N_s \times T_r}$, where the index ``$0$'' selects the charging demand feature. An analogous procedure is followed to construct $\mathbf{H}_{t,\mathrm{demd}}^\mathrm{wek}$ using $(\mathbf{X}^\mathrm{wek}_t)_{::0}$.

Below details the hypergraph construction process for the recent timescale as an example. This process first computes a pairwise similarity matrix, denoted as $\mathbf{\Omega}_t^{\mathrm{rec}} \in \mathbb{R}^{N_s \times N_s}$, for the recent demand using Pearson correlation. Spectral clustering is then performed on the derived similarity matrix to identify $K$ hyperedges. The clustering calculates the graph Laplacian matrix, expressed as $\mathbf{L}_t^{\mathrm{rec}} = \mathbf{D}_{t}^{\mathrm{rec}} - \mathbf{\Omega}_t^{\mathrm{rec}}$, where $(\mathbf{D}_{t}^{\mathrm{rec}})_{pp} = \sum_{p'=1}^{N_s} (\mathbf{\Omega}_t^{\mathrm{rec}})_{pp'}$ is the degree matrix. The resulted eigenvectors of the Laplacian, denoted as $\mathbf{e}_1, \dots, \mathbf{e}_{K}$, are stacked to form the matrix $\mathbf{EigVec}_t^{\mathrm{rec}} \in \mathbb{R}^{N_s \times K}$. Soft assignments of stations to hyperedges are then applied as
\begin{equation}
\label{eq:spectral_clustering}
    (\mathbf{H}_{t,\mathrm{demd}}^{\mathrm{rec}})_{pk} = \frac{|(\mathbf{EigVec}_t^{\mathrm{rec}})_{pk}|}{\sum_{k'=1}^{K} |(\mathbf{EigVec}_t^{\mathrm{rec}})_{pk'}|},
\end{equation}
resulting in the normalized, non-negative incidence matrix $\mathbf{H}_{t,\mathrm{demd}}^{\mathrm{rec}} \in [0,1]^{N_s \times K}$. Note that the input demand series $(\mathbf{X}^\mathrm{rec}_t)_{::0}$ is batch-dependent (i.e., different for each item in a batch), this hypergraph construction is thus performed independently for each batch item, generating batch-wise hypergraph incidence matrix $\mathbf{H}_{t,\mathrm{demd}}^\mathrm{rec} \in \mathbb{R}^{B \times N_s \times K}$, where $B$ represents the batch size.

A significant advantage of our hypergraph construction methodology in HyperCast is its consistent use of soft assignments for defining the relationships between charging stations and hyperedges, across both the distance-based and demand-based views.

In the distance-based hypergraph, FCM naturally produces soft memberships which allow a charging station to belong to multiple geographically-defined hyperedges to varying degrees. This is particularly beneficial since real-world spatial clusters are often not sharply delineated; a station might be situated in a region that reasonably shares characteristics of several nearby zones. Soft assignments capture this geographical ambiguity and overlap more faithfully than rigid assignments.

Similarly, for the demand-based hypergraphs, the normalization of eigenvector components from spectral clustering provides soft assignments, enabling a station to be associated with multiple hyperedges that represent distinct groups of demand behaviors. This is crucial as a station may present multifaceted demand patterns, sharing some characteristics with one group (e.g., weekday commuter peaks) and other traits with another (e.g., weekend leisure patterns).

In both views, therefore, the use of soft memberships provides a more flexible and realistic representation of how stations relate to different groupings, either geographical or functional. Instead of forcing an exclusive membership, this approach acknowledges that stations can have partial affiliations or embody transitional characteristics. The richer representation of inter-station relationships is a key contributor to the model's ability to learn and predict complex spatiotemporal charging patterns effectively.

\begin{figure*}[!t]
    \centering
    \includegraphics[width=.80\linewidth]{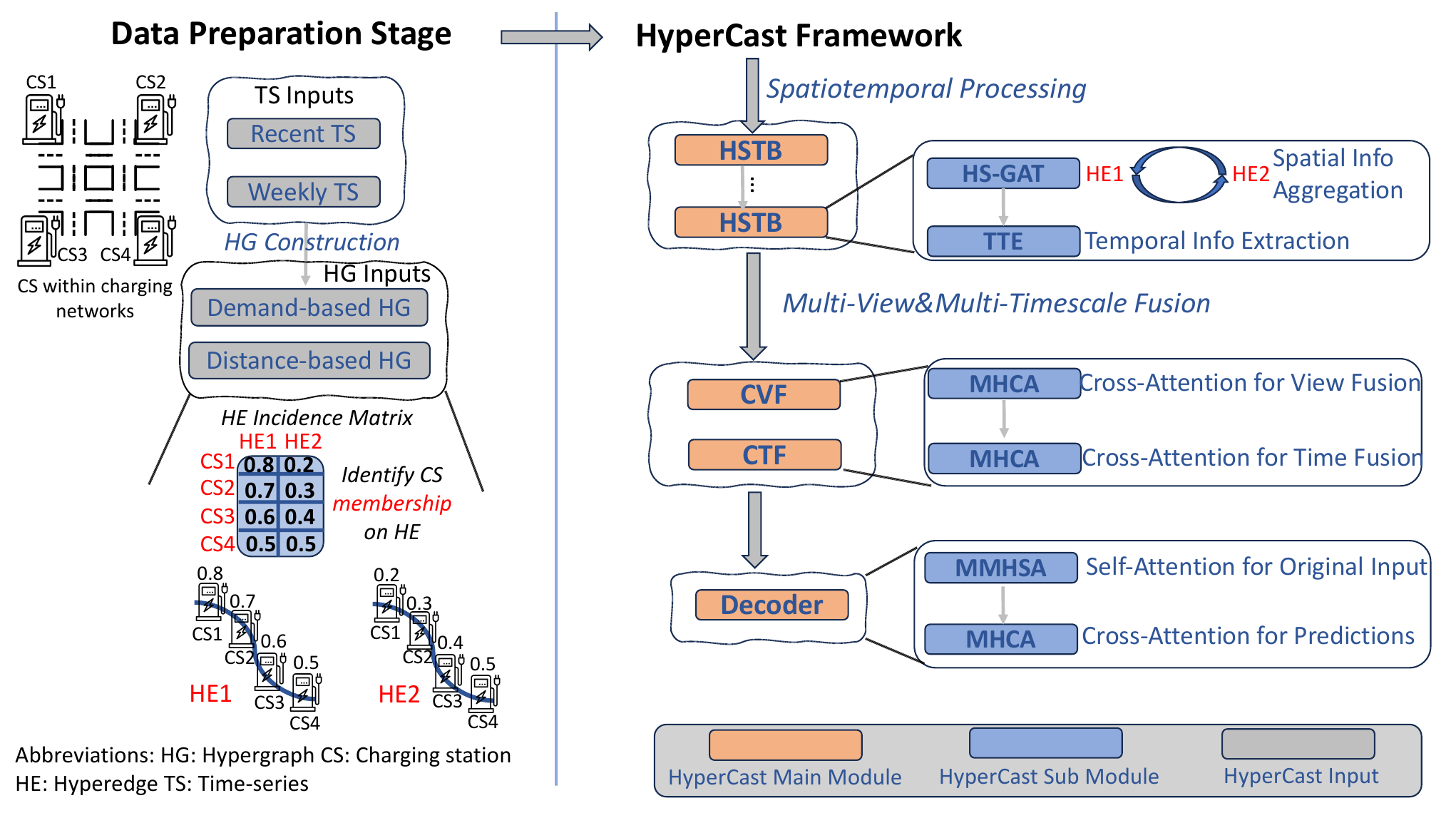}
    \caption{Illustrative diagram of HyperCast.}
    \label{fig:hypercast}
\end{figure*}

\subsection{Hyper-Spatiotemporal Block} \label{subsec:method_HSTB}

The \textit{Hyper-Spatiotemporal Block} (HSTB) is a core component of HyperCast, designed to jointly learn spatial and temporal representations from the input data. Note that the combination of multiple timescales and hypergraph views creates four processing streams, characterized by unique pairings of timescale (denoted as $u \in \{\mathrm{rec, wek}\}$) and hypergraph view (denoted as $v \in \{\mathrm{dist, demd}\}$). Each of the four processing streams passes through a sequence of HSTBs. We define the number of HSTBs as $L_\mathrm{HSTB}$. 

Before fed into the HSTBs, their raw features are firstly projected into unified latent space via a neural linear layer denoted as $\mathrm{Linear}(\cdot)$. Let $T_\mathrm{seq}^u$ represent the sequence length for the timescale $u$: $T_r$ if $u=\mathrm{rec}$ or $T_w$ if $u=\mathrm{wek}$. Let $t' \in \{1, \dots, T_{\mathrm{seq}}^u\}$ be the time step index within the respective sequence length, $d_h$ denote the feature embedding dimension. The projection process can be formulated as
\begin{equation}
\label{eq:init_embed}
    (\mathbf{Z}_{t,(0)}^{u})_{bpt'} = \mathrm{Linear}((\mathbf{X}^{u}_t)_{bpt'}) \in \mathbb{R}^{B\times N_s \times T_\mathrm{seq}^u \times d_h}.
\end{equation}

Given timescale $u$ and view $v$, we define the input to the $l$-th HSTB (i.e., the output of the previous HSTB) as  $\mathbf{Z}_{t,(l-1)}^{u,v}$, where $l$ is the HSTB index within the set of $\{1,\cdots, L_\mathrm{HSTB}\}$. The input of the first block, i.e., $\mathbf{Z}_{t,(0)}^{u,v}$, is effectively the initial embedding $\mathbf{Z}_{t,(0)}^{u}$ defined in Eq. \eqref{eq:init_embed}, which means that $\mathbf{Z}_{t,(0)}^{u,\mathrm{dist}}$ and $\mathbf{Z}_{t,(0)}^{u,\mathrm{demd}}$ are identical. The HSTB input goes through two main sub-modules: a \textit{Hyper-Spatial GAT} (HS-GAT) and a \textit{Temporal Transformer Encoder} (TTE).

\subsubsection{HS-GAT} HS-GAT aims to capture and process spatial dependencies by utilizing constructed hypergraphs. Here, we further denote the hypergraph incidence matrix as $\mathbf{H}^{u}_{t,v}$: $\mathbf{H}_\mathrm{dist}$ if view $v=\mathrm{dist}$, or $\mathbf{H}_{t,\mathrm{demd}}^{u}$ if view $v=\mathrm{demd}$. 

The HS-GAT operation begins with \textit{Node-to-Hyperedge Aggregation}. For each batch $b$ and historical step $t'$, we define features aggregated on hyperedge as $(\mathbf{F}_{t,(l)})_{bt'} \in \mathbb{R}^{K \times d_h}$, which are computed by projecting node features onto hyperedges through the hyperedge incidence matrix:
\begin{equation}
\label{eq:hyperedge_feature_agg}
    (\mathbf{F}_{t,(l)})_{bt'} = ((\mathbf{H}^{u}_{t,v})_b)^T (\mathbf{Z}_{t,(l-1)}^{u,v})_{bt'},
\end{equation}
where the above collection forms $\mathbf{F}_{t,(l)} \in \mathbb{R}^{B \times T_{\mathrm{seq}}^u \times K \times d_h}$.

Next, \textit{Hyperedge Graph Attention} is applied. A graph attention network (GAT)~\cite{gat} operates on the aggregated features of each hyperedge, i.e., $(\mathbf{F}_{t,(l)})_{bt'k} \in \mathbb{R}^{d_h}$. Let $N_h^\mathrm{gat}$ denote the number of attention heads, thereby making the head dimension $d_\mathrm{head}^\mathrm{gat}$ equal to $d_h/N_h^\mathrm{gat}$. For each attention head, a linear transformation is firstly applied as
\begin{equation}
\label{eq:gat_proj}
    \mathbf{f}_{bt'k}^{(h)} = \mathbf{W}_\mathrm{gat}^{(h)} (\mathbf{F}_{t,(l)})_{bt'k} \in \mathbb{R}^{d_\mathrm{head}^\mathrm{gat}},
\end{equation}
where $\mathbf{W}_\mathrm{gat}^{(h)}$ 
is a learnable weight matrix for the $h$-th head, projecting features to a $d_\mathrm{head}^\mathrm{gat}$-dimensional space.

Attention coefficients $e_{bt'kj}^{(h)}$ between hyperedge $k$ and hyperedge $k'$ within the $k$-th hyperedge's neighborhood $\mathcal{N}_k$ are then computed as
\begin{equation}
\label{eq:gat_attn_logits}
    e_{bt'kk'}^{(h)} = \mathrm{LeakyReLU}( (\mathbf{a}_\mathrm{gat}^{(h)})^T [\mathbf{f}_{bt'k}^{(h)} || \mathbf{f}_{bt'k'}^{(h)}] ),
\end{equation}
where $\mathrm{LeakyReLU}(\cdot)$ is the leaky rectified linear unit activation function, $||$ denotes concatenation, and $\mathbf{a}_\mathrm{gat}^{(h)} \in \mathbb{R}^{2 d_\mathrm{head}^\mathrm{gat}}$ are learnable parameters. These coefficients are normalized using the softmax function as
\begin{equation}
\label{eq:gat_attn_weight}
    \alpha_{bt'kk'}^{(h)} = \mathrm{softmax}_{k'\in \mathcal{N}_k}(e_{bt'kk'}^{(h)}).
\end{equation}
The attention-weighted aggregation per head then produces the updated hyperedge features of the specific head, formulated as
\begin{equation}
\label{eq:gat_update_hyperedge_feature}
    (\mathbf{F'}^{(h)}_{t,(l)})_{bt'k} = \sum_{k' \in \mathcal{N}_k} \alpha_{bt'kk'}^{(h)} \mathbf{f}_{bt'k'}^{(h)} \in \mathbb{R}^{d_\mathrm{head}^\mathrm{gat}}.
\end{equation}
The outputs of all heads are further concatenated as
\begin{equation}
\label{eq:gat_update_hyperedge_feature_concat}
    \begin{aligned}
        (\mathbf{F}'_{t,(l)})_{bt'k} &= [(\mathbf{F}'^{(1)}_{t,(l)})_{bt'k}, ||  \cdots, || (\mathbf{F}'^{(N_h^\mathrm{gat})}_{t,(l)})_{bt'k}].
    \end{aligned}
\end{equation}
The above collected outputs form $\mathbf{F}'_{t,(l)} \in \mathbb{R}^{B \times T_{\mathrm{seq}}^u \times K \times d_h}$.

At the end of HS-GAT, \textit{Hyperedge-to-Node Projection} maps the updated hyperedge features back to the node domain via the hyperedge incidence matrix, which can be formulated as
\begin{equation}
\label{eq:gat_update_node_feature}
    ({\mathbf{Z}'}_{t,(l-1)}^{u,v})_{bt'} = (\mathbf{H}^{u}_{t,v})_b (\mathbf{F}'_{t,(l)})_{bt'},
\end{equation}
yielding updated node features ${\mathbf{Z}'}_{t,(l-1)}^{u,v} \in \mathbb{R}^{B \times N_s \times T_{\mathrm{seq}}^u \times d_h}$. 

Finally, a \textit{Residual Connection} is applied and termed $\mathrm{RC}(\cdot)$, which consists of layer normalization (denoted as $\mathrm{LN}(\cdot)$) and Dropout (denoteds as $\mathrm{Drop}(\cdot)$). This operation is detailed as
\begin{equation}
    \hspace{-2.5mm} ({\mathbf{Z}'}_{t,(l-1)}^{u,v})_{bt'} \leftarrow \mathrm{LN}( (\mathbf{Z}_{t,(l-1)}^{u,v})_{bt'}  + \mathrm{Drop}(({\mathbf{Z}'}_{t,(l-1)}^{u,v})_{bt'}) ),
\end{equation}
further simplified as $\mathrm{RC}((\mathbf{Z}_{t,(l-1)}^{u,v})_{bt'}, ({\mathbf{Z}'}_{t,(l-1)}^{u,v})_{bt'})$. Here, we use the left-arrow ``$\leftarrow$'' to represent variable updates for notation brevity. The output of the HS-GAT is written as ${\mathbf{Z}'}_{t,(l-1)}^{u,v} \in \mathbb{R}^{B \times N_s \times T_{\mathrm{seq}}^u \times d_h}$.

\subsubsection{TTE}: TTE captures and models temporal dependencies within the sequences processed by the HS-GAT. For each station $p$ in batch $b$, its input sequence $({\mathbf{Z}'}_{t,(l-1)}^{u,v})_{bp} \in \mathbb{R}^{T_{\mathrm{seq}}^u \times d_h}$ is processed by a \textit{Standard Transformer Encoder Layer} (STEL). The STEL first employs the \textit{Multi-Head Self-Attention (MHSA)}, which generates query, key, and value matrices from $({\mathbf{Z}'}_{t,(l-1)}^{u,v})_{bp}$ through linear projection. 

Let $N_h^\mathrm{MHSA}$ denote the number of MHSA's attention heads, each with dimension $d_\mathrm{head}^\mathrm{MHSA} = d_h / N_h^\mathrm{MHSA}$. For each head $h$, the input projection process is formulated as
\begin{align}
    \mathbf{Q}^{(h)}_{t,(l)} &= ({\mathbf{Z}'}_{t,(l-1)}^{u,v})_{bp} \mathbf{W}_Q^{(h)} \in \mathbb{R}^{T_{\mathrm{seq}}^u \times d_\mathrm{head}^\mathrm{MHSA}}, \\
    \mathbf{K}^{(h)}_{t,(l)} &= ({\mathbf{Z}'}_{t,(l-1)}^{u,v})_{bp} \mathbf{W}_K^{(h)} \in \mathbb{R}^{T_{\mathrm{seq}}^u \times d_\mathrm{head}^\mathrm{MHSA}}, \\
    \mathbf{V}^{(h)}_{t,(l)} &= ({\mathbf{Z}'}_{t,(l-1)}^{u,v})_{bp} \mathbf{W}_V^{(h)} \in \mathbb{R}^{T_{\mathrm{seq}}^u \times d_\mathrm{head}^\mathrm{MHSA}},
\end{align}
using learnable weight matrices $\mathbf{W}_Q^{(h)}, \mathbf{W}_K^{(h)}, \mathbf{W}_V^{(h)}$. 

Self-attention weights in the temporal domain are then calculated based on the query and key matrices as
\begin{equation}
\label{eq:HSTB_temp_attn_weight}
    \mathbf{A}^{(h)}_{t,(l)} = \mathrm{softmax}(\frac{\mathbf{Q}^{(h)}_{t,(l)} (\mathbf{K}^{(h)}_{t,(l)})^T}{\sqrt{d_\mathrm{head}^\mathrm{MHSA}}}) \in \mathbb{R}^{T_{\mathrm{seq}}^u \times T_{\mathrm{seq}}^u}.
\end{equation}
Similar to HS-GAT, each head's output, derived by $\mathbf{A}^{(h)}_{t,(l)} \mathbf{V}^{(h)}_{t,(l)}  \in \mathbb{R}^{T_{\mathrm{seq}}^u \times d_\mathrm{head}^\mathrm{MHSA}}$, is then concatenated with results fed into the final linear projection, leading to the MHSA's output defined as $\mathrm{MHSA}(({\mathbf{Z}'}_{t,(l-1)}^{u,v})_{bp}) \in \mathbb{R}^{T_{\mathrm{seq}}^u \times d_h}$. The MHSA is further followed by a residual connection module, written as 
\begin{equation}
\label{eq:res_conn_after_MHSA}
    \hspace{-3.5mm} ({\mathbf{Z}'}_{t,(l-1)}^{u,v})_{bp} \leftarrow \mathrm{RC}(({\mathbf{Z}'}_{t,(l-1)}^{u,v})_{bp}, \mathrm{MHSA}(({\mathbf{Z}'}_{t,(l-1)}^{u,v})_{bp})).
\end{equation}

Finally, a position-wise feed-forward network (FFN), consisting of two linear neural layers with an activation of the rectified linear unit, is applied and followed by another residual connection module, which generates the output of the $l$-th HSTB. Such a process can be formulated as
\begin{equation}
\label{eq:res_conn_after_FFN}
    ({\mathbf{Z}}_{t,(l)}^{u,v})_{bp} = \mathrm{RC}(({\mathbf{Z}'}_{t,(l-1)}^{u,v})_{bp}, \mathrm{FFN}( ({\mathbf{Z}'}_{t,(l-1)}^{u,v})_{bp}) )).
\end{equation}

In summary, we denote the entire STEL operation as $({\mathbf{Z}}_{t,(l)}^{u,v})_{bp} = \mathrm{STEL} (({\mathbf{Z}'}_{t,(l-1)}^{u,v})_{bp})$. Reassembling outputs for all stations and batches leads to the output of the $l$-th HSTB for stream $(u,v)$, denoted as $\mathbf{Z}_{t,(l)}^{u,v} \in \mathbb{R}^{B \times N_s \times T_{\mathrm{seq}}^u \times d_h}$.

After going through all HSTBs, we obtain four outputs:
$ \mathbf{Z}_{t,(L_\mathrm{HSTB})}^{\mathrm{rec,dist}}$,
$\mathbf{Z}_{t,(L_\mathrm{HSTB})}^{\mathrm{rec,demd}}$,
$\mathbf{Z}_{t,(L_\mathrm{HSTB})}^{\mathrm{wek,dist}}$, and
$\mathbf{Z}_{t,(L_\mathrm{HSTB})}^{\mathrm{wek,demd}} $. For brevity, these are further rewritten as $\mathbf{Z}_{t}^{\mathrm{rec},\mathrm{dist}}$, $\mathbf{Z}_{t}^{\mathrm{rec},\mathrm{demd}}$, $\mathbf{Z}_{t}^{\mathrm{wek},\mathrm{dist}}$, and $\mathbf{Z}_{t}^{\mathrm{wek},\mathrm{demd}}$, respectively.

\subsection{Cross-View Fusion} \label{subsec:method_CVF}

The \textit{Cross-View Fusion} (CVF) module is responsible for integrating the representations learned from the two different hypergraph views for each timescale. 

The inputs to the CVF module are denoted as $\mathbf{Z}_{t}^{u,\mathrm{dist}}$ and $\mathbf{Z}_{t}^{u,\mathrm{demd}}$, which are then stacked along a new dimension to form $\mathbf{Z}_t^\mathrm{u} \in \mathbb{R}^{B \times N_s \times 2 \times T_{\mathrm{seq}}^u \times d_h}$, where ``$2$'' represents the number of views. The fusion mechanism first reshapes the input $\mathbf{Z}_t^\mathrm{u}$ from $(B, N_s, 2, T_{\mathrm{seq}}^u, d_h)$ to $(B, N_s, T_{\mathrm{seq}}^u, 2, d_h)$, and then further to $(B \cdot N_s \cdot T_{\mathrm{seq}}^u, 2, d_h)$. This reshaping operation creates a batch of sequences, each of which has two views.

Let $(\mathbf{Z}_t^u)_{bpt'}$ denote one such sequence which undergoes an initial linear embedding, formulated as $(\mathbf{Z}_t^u)_{bpt'} \leftarrow \mathrm{Linear}((\mathbf{Z}_t^u)_{bpt'})$. The embedded sequences are then processed by an STEL, with results written as $(\mathbf{Z}_t^u)_{bpt'} \leftarrow \mathrm{STEL} ((\mathbf{Z}_t^u)_{bpt'})$.

Given the output of the STEL, a single $d_h$-dimensional vector is extracted to represent the fused information by selecting the features corresponding to the last view in the input sequence. This selection process is expressed as
\begin{equation}
\label{eq:view_fusion_selection}
    (\mathbf{Z}_t^{u,\mathrm{CVF}})_{bpt'} = (\mathbf{Z}_t^u)_{bpt'1},
\end{equation}
which is subsequently passed through a final linear projection layer. The resulting $d_h$-dimensional fused vectors are reshaped back to form the CVF's output $\mathbf{Z}_t^{u,\mathrm{CVF}} \in \mathbb{R}^{B \times N_s \times T_{\mathrm{seq}}^u \times d_h}$. Compared to simpler fusion methods, such as concatenation and element-wise averaging, our CVF stands out for creating a more dynamic and contextualized fusion by effectively utilizing the two views to interact and inform one another.

\subsection{Cross-Timescale Fusion} \label{subsec:method_CTF}

The \textit{Cross-Timescale Fusion} (CTF) module integrates information from the two different timescales following their processing by the CVF module. The CTF takes $\mathbf{Z}_t^{\mathrm{rec},\mathrm{CVF}}$ and $\mathbf{Z}_t^{\mathrm{wek},\mathrm{CVF}}$ as inputs and uses the \textit{Multi-Head Cross-Attention} (MHCA) to fuse the two timescales.

Let $N_\mathrm{h}^\mathrm{MHCA}$ denote the number of MHCA's attention heads, with the head dimension of $d_\mathrm{head}^\mathrm{MHCA} = d_h / N_h^\mathrm{MHCA}$. For each batch $b$ and station $p$, the query, key, and value matrices of the MHCA are derived using learnable weight matrices as
\begin{align}
\label{eq:MHCA_QKV}
    \mathbf{Q}^{(h)}_t &= (\mathbf{Z}_t^{\mathrm{rec},\mathrm{CVF}})_{bp} \mathbf{W}_Q^{(h)} \in \mathbb{R}^{T_r \times d_\mathrm{head}^\mathrm{MHCA}}, \\
    \mathbf{K}^{(h)}_t &= (\mathbf{Z}_t^{\mathrm{wek},\mathrm{CVF}})_{bp} \mathbf{W}_K^{(h)} \in \mathbb{R}^{T_w \times d_\mathrm{head}^\mathrm{MHCA}}, \\
    \mathbf{V}^{(h)}_t &= (\mathbf{Z}_t^{\mathrm{wek},\mathrm{CVF}})_{bp} \mathbf{W}_V^{(h)} \in \mathbb{R}^{T_w \times d_\mathrm{head}^\mathrm{MHCA}}.
\end{align}
Attention weights are then computed as 
\begin{equation}
\label{eq:MHCA_attn}
    \mathbf{A}^{(h)}_t = \mathrm{softmax}(\frac{\mathbf{Q}^{(h)}_t (\mathbf{K}^{(h)}_t)^T}{\sqrt{d_\mathrm{head}^\mathrm{MHCA}}}) \in \mathbb{R}^{T_r \times T_w}.
\end{equation}
The above asymmetric design aims to exploit the distinct roles of the two timescales, where the recent window holds the most up-to-date information and the weekly window provides historical context that can explain or modulate the current state. Therefore, the results are anchored in the recent but informed by the weekly patterns. The outputs from all heads are concatenated:
\begin{equation}
\label{eq:MHCA_head_concat}
    (\mathbf{Z}_t^\mathrm{CTF})_{bp} = [\mathbf{A}^{(1)}_t \mathbf{V}^{(1)}_t, || \cdots, || \mathbf{A}^{(N_h^\mathrm{MHCA})}_t \mathbf{V}^{(N_h^\mathrm{MHCA})}_t ],
\end{equation}
followed by a final linear projection.

This entire MHCA process is denoted as $(\mathbf{Z}_t^\mathrm{CTF})_{bp} = \mathrm{MHCA} (\mathbf{Z}_t^{\mathrm{rec},\mathrm{CVF}}, \mathbf{Z}_t^{\mathrm{wek},\mathrm{CVF}})$.
A residual connection then combines this output with the recent timescale representation:
\begin{equation}
    (\mathbf{Z}_t^\mathrm{CTF})_{bp} \leftarrow \mathrm{RC}(\mathbf{Z}_t^{\mathrm{rec},\mathrm{CVF}}, (\mathbf{Z}_t^\mathrm{CTF})_{bp}).
\end{equation}
The resulting $\mathbf{Z}_t^\mathrm{CTF} \in \mathbb{R}^{B \times N_s \times T_r \times d_h}$ is considered the extracted spatiotemporal representation. 

\subsection{Forecasting Decoder} \label{subsec:method_decoder}

The transformer decoder uses the fused spatiotemporal representation to generate the final predictions.

Let $(\mathbf{Z}_{t,\mathrm{tgt}}^{(0)})_{bp}$ denote the initialized learnable input target sequence of the decoder and $N_\mathrm{dec}$ be the number of decoder layers. The input of the $l$-th decoder layer can be thus denoted as $(\mathbf{Z}_{t,\mathrm{tgt}}^{(l-1)})_{bp}$.

Each decoder layer first applies the \textit{Masked Multi-Head Self-Attention (MMHSA)} to the target sequence. The MMHSA is similar to MHSA but prevents positions from attending to subsequent positions by setting attention weights $(\mathbf{A}^{(h)}_t)_{ij}$ to $0$ if $i > j$. The MMHSA operator is formulated as
\begin{equation}
    (\mathbf{Z'}_{t,\mathrm{tgt}}^{(l-1)})_{bp} = \mathrm{MMHSA}((\mathbf{Z}_{t,\mathrm{tgt}}^{(l-1)})_{bp}).
\end{equation}
The MMHSA is followed by an MHCA, where the query is the output from the MMHSA and the key and value are spatiotemporal information derived from the CTF, i.e., $(\mathbf{Z}_t^\mathrm{CTF})_{bp}$. The output of the $l$-th decoder layer is written as
\begin{equation}
    (\mathbf{Z}_{t,\mathrm{tgt}}^{(l)})_{bp} = \mathrm{MHCA} ((\mathbf{Z'}_{t,\mathrm{tgt}}^{(l-1)})_{bp}, (\mathbf{Z}_t^\mathrm{CTF})_{bp}).
\end{equation}
Note that both MMHSA and MHCA steps are followed by residual connections. The final output of the decoder can be written as $\mathbf{Z}_{t,\mathrm{tgt}}^{(N_\mathrm{dec})} \in \mathbb{R}^{(B \cdot N_s) \times T_f \times d_h}$, which is then passed through a linear projection layer that maps the $d_h$-dimensional features to a single predicted value for each time step:
$\mathbf{\hat{Y}}_t^\mathrm{flat} = \mathrm{Linear}(\mathbf{Z}_{t,\mathrm{tgt}}^{(N_\mathrm{dec})}) \in \mathbb{R}^{(B \cdot N_s) \times T_f \times 1}$.
Finally, this flat output is reshaped to derive the forecast output $\mathbf{\hat{Y}}_t \in \mathbb{R}^{B \times N_s \times T_f}$.

In summary, benefiting from multi-view hypergraph construction and multi-timescale time-series input design, the HyperCast framework presents significant advantages in modeling how charging stations relate, capturing both geographical ambiguities and multifaceted demand characteristics that simpler models might overlook. By systematically processing these rich representations through tailored HSTBs and progressively fusing information across different views and timescales using self-attention and cross-attention mechanisms, HyperCast is adept at disentangling and then synergizing the complex, interwoven spatial and temporal factors that drive EV charging demand. The HyperCast's workflow is depicted in Fig. \ref{fig:hypercast}.

\begin{remark}[Computation Complexity Analysis]
    The computation complexity of HyperCast can be analyzed from its core components: the HSTB, the CVF, the CTF, the forecasting decoder. For the CVF and CTF, they are comparatively lightweight due to their single-layer structure designed for information fusion. Hence, our analysis focuses on the HSTB and the forecasting decoder.

    Within each HSTB, the HS-GAT operation requires both node-to-hyperedge and hyperedge-to-node projections, each with a cost of $O(N_sKd_h)$, together with hyperedge-level attention of complexity $O(K^2d_hN_h^\mathrm{gat})$. The following TTE applied to each station has a complexity of $O(BN_sT_\mathrm{seq}^2d_h)$. Given these HSTBs are stacked $L_\mathrm{HSTB}$ times, the overall spatiotemporal processing cost grows linearly in $L_\mathrm{HSTB}$, written as $ O(L_\mathrm{HSTB} (N_sKd_h + K^2d_hN_h^\mathrm{gat} + BN_sT_\mathrm{seq}^2d_h))$.

    The forecasting decoder contributes additional cost dominated by the quadratic self-attention term $O(BN_sT_f^2d_h)$ from the masked multi-head self-attention and the cross-attention term $O(BN_sT_fT_rd_h)$, repeated across $N_\mathrm{dec}$ decoder layers. The computation cost of the decoder can be written as $O(N_\mathrm{dec} B N_s (T_f^2 + T_fT_r) d_h)$.
\end{remark}

\begin{table}[!t]
\centering
\caption{Forecasting method evaluation under different forecast horizons in the Palo Alto dataset.}
\resizebox{\columnwidth}{!}{
\begin{tabular}{c cc cc cc }

\toprule

\multirow{2}{*}{%
  \diagbox[height=2\normalbaselineskip]{Method}{Metric}%
}  & \multicolumn{2}{c}{$\mathrm{MSE}$} & \multicolumn{2}{c}{$\mathrm{MAE}$ (kW)} & \multicolumn{2}{c}{$\mathrm{R^2}$} \\

\cmidrule(lr){2-3} \cmidrule(lr){4-5} \cmidrule(lr){6-7} 

 & \multicolumn{1}{c}{$T_f=3$} & \multicolumn{1}{c}{$T_f=7$} & \multicolumn{1}{c}{$T_f=3$} & \multicolumn{1}{c}{$T_f=7$} & \multicolumn{1}{c}{$T_f=3$} & \multicolumn{1}{c}{$T_f=7$} \\

\midrule



LSTM~\cite{Adil2024_lstm_forecasting} & $1516$ & $1499$ & $28.9$ & $28.6$ & $0.54$ & $0.52$ \\

GRU & $1497$ & $1446$ & $28.7$ & $28.1$ & $0.54$ & $0.52$ \\

TCN~\cite{tcn} & $1400$ & $1484$ & $27.3$ & $28.0$ & $0.54$ & $0.52$  \\

DeepAR~\cite{deepar} & $13416$ & $13520$ & $97.5$ & $98.0$ & $-2.43$ & $-2.46$ \\

NBEATS~\cite{nbeats} & $1358$ & $1438$ & $27.4$ & $28.1$ & $0.65$ & $0.63$ \\

Informer~\cite{informer} & $12508$ & $4318$  & $92.7$ & $48.7$  & $-2.20$ & $-0.10$ \\

Autoformer~\cite{autoformer} & $1469$ & $1540$ & $28.3$ & $29.2$  & $0.62$  & $0.61$ \\

FEDformer~\cite{fedformer} & $1410$ & $1451$ & $27.7$ & $28.0$ &  $0.64$ & $0.62$ \\

DLinear~\cite{dlinear} & $3312$  & $4318$  & $42.4$ & $48.7$  & $0.15$ & $-0.10$ \\

NHITS~\cite{nhits} & $1556$  & $1771$ & $29.5$ & $31.3$ & $0.60$ & $0.54$ \\

\midrule\midrule

GCN~\cite{Li2021_gcn_prediction} & $1593$ & $1619$  & $29.8$  & $30.1$ & $0.59$  & $0.58$ \\

GAT~\cite{Li2024_att_lstm_with_dn_update} & $1616$ & $1755$ & $30.1$ & $34.0$  & $0.58$ & $0.55$ \\

T-GCN~\cite{Hüttel2021_TGCN_forecast} & $1498$ & $1540$ & $29.4$ & $29.2$ & $0.63$ & $0.61$ \\

GWN~\cite{gwn} & $1404$  & $1401$  & $28.2$ & $28.1$ & $0.65$ & $0.64$ \\

ASTGCN~\cite{Shi2024_ASTGCN_forecast} & $1330$ & $1488$ & $27.2$  & $28.3$  & $0.66$  & $0.62$ \\

\midrule\midrule

HyperGCN-Dist~\cite{hypergcn} & $1075$ & $990$ & $20.2$ & $19.5$ & $0.82$ & $0.82$ \\ 

HyperGCN-Demd~\cite{hypergcn} & $924$ & $997$ & $23.2$ & $22.6$ & $0.77$ & $0.78$ \\ 

\midrule\midrule

$\textbf{Ours -- } \mathrm{Max.}$ & $\bm{808}$ & $\bm{830}$ & $\bm{21.3}$ & $\bm{21.8}$ & $\bm{0.89}$ & $\bm{0.90}$ \\ 

$\textbf{Ours -- } \mathrm{Min.}$ & $\bm{416}$ & $\bm{416}$ & $\bm{14.8}$ & $\bm{15.0}$ & $\bm{0.80}$ & $\bm{0.79}$ \\

\bottomrule
\end{tabular}
}

\label{tab:benchmark_comp}
\end{table} 

\begin{figure}[!t]
    \centering
    \includegraphics[width=.8\linewidth]{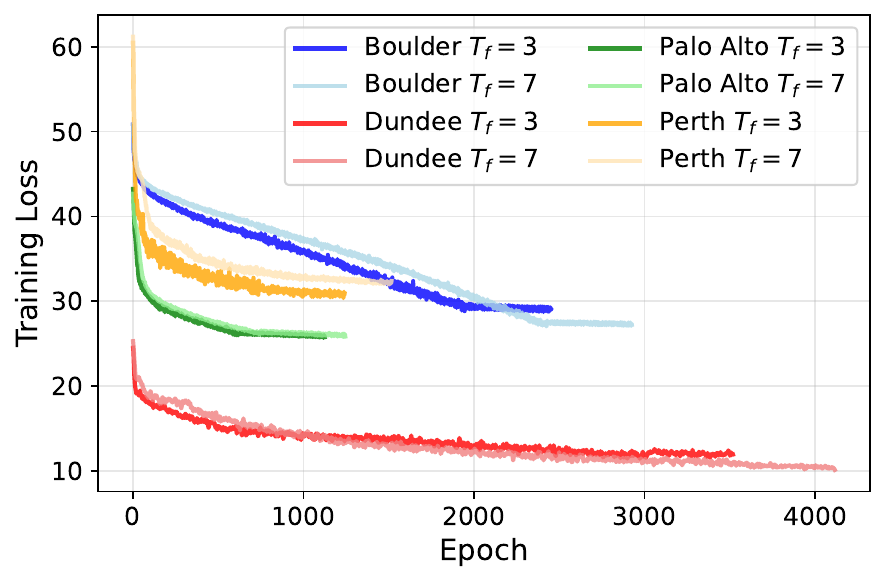}
    \caption{Training loss curves across four datasets and two forecasting horizons.}
    \label{fig:train_loss_curves}
\end{figure}

\section{Experimental Results} \label{sec:exp}

\subsection{Experiment Settings}

\subsubsection{Datasets}
We use four widely-adopted and publicly-accessible EV charging datasets, which are released by the city of Palo Alto, U.S.~\cite{dataCityofPaloAlto}, the city of Perth, Scotland~\cite{dataCityofPerth}, the city of Boulder, U.S.~\cite{dataCityofBoulder}, and the city of Dundee, Scotland~\cite{dataCityofDundee}. Missing values in these datasets are imputed using the $\texttt{imputeTS}$ package~\cite{imputets}. For model training and evaluation, the datasets are split into training and test sets with a ratio of $80:20$. 

\subsubsection{Hyperparameters}

The look-back horizons for the recent and the weekly time-scales, i.e., $T_r$ and $T_w$, are selected from $\{ 7, 14, 21, 28 \}$ and $\{ 1, 2, 3 ,4 \}$, respectively. The forecast horizon, denoted as $T_f$, ranges over $\{3, 7\}$. The total number of features in the raw input, denoted as $F_\mathrm{raw}$, is set to $7$, including charging demand and calendar feature information. Note that our HyperCast framework is adaptable to richer features, such as weather and traffic.

The number of charging stations, i.e., $N_s$, are $6$, $26$, $6$, and $10$ for Boulder, Dundee, Palo Alto, and Perth, respectively. The number of hyperedges, i.e., $K$, varies across datasets due to their different number of charging stations. This is because $K$ represents the number of groups that we aim to identify within the charging network. Hence, the number of groups cannot conceptually exceed the number of stations being grouped. Thus, the upper bound for $K$ is set as $N_s-1$. Specifically, for Boulder, the range of $K$ is $\{ 2, 3, 4 ,5 \}$; for Dundee, $\{ 5, 10, 15, 20, 25 \}$; for Palo Alto, $\{ 1, 2, 3, 4, 5 \}$; and for Perth, $\{ 1, 3, 6, 9 \}$. These choices aim to span a range up to the maximum number of charging stations in each dataset.

The HyperCast model is configured with a batch size $B$ of $32$, a hidden dimension $d_h$ of $64$, the number of HSTB $L_\mathrm{HSTB}$ as $3$, the number of attention heads $N_h^\mathrm{gat}$, $N_h^\mathrm{MHSA}$, $N_h^\mathrm{MHCA}$ all as $8$, the number of decoder layer $N_\mathrm{dec}$ as $2$. The dropout ratio is set to $0.1$.

The training process employs the Adam optimizer to minimize the mean squared error (MSE) loss function. The learning rate is initialized at $10^{-5}$ and is managed by a scheduler that can reduce it to a minimum of $10^{-6}$. An early stopping mechanism is implemented: training terminates if the training loss does not show significant improvement (i.e., at least absolute value of $1.0$) for $500$ consecutive epochs. The maximum number of training epochs is set to $4500$. Model training is performed on one NVIDIA V100 GPU. The Python and PyTorch versions used are 3.10.13 and 1.21.0, respectively. The training loss curves across four datasets and two forecasting horizons are depicted in Fig. \mbox{\ref{fig:train_loss_curves}}, showing that HyperCast training ends before the maximum training epoch due to the incorporated early stopping mechanism.

\subsubsection{Baselines and Evaluation Metrics}

The performance of HyperCast is compared against two categories of baseline models: (i) sequential-based methods, including common models such as long short-term memory (LSTM), gated recurrent units (GRU), and temporal convolution network (TCN) \cite{tcn}, as well as state-of-the-art models like DeepAR \cite{deepar}, NBEATS \cite{nbeats}, Informer \cite{informer}, Autoformer \cite{autoformer}, FEDformer \cite{fedformer}, DLinear \cite{dlinear}, and NHITS \cite{nhits}. These models are implemented using the $\texttt{NeuralForecast}$ package \cite{neuralforecast}; (ii) graph-based methods, including common graph-based models like GCN \cite{Li2021_gcn_prediction} and GAT \cite{Li2024_att_lstm_with_dn_update}, spatio-temporal models such as T-GCN \cite{Hüttel2021_TGCN_forecast} and GWN \cite{gwn}, and the state-of-the-art ASTGCN \cite{Shi2024_ASTGCN_forecast}; and 3) hypergraph-based methods -- two variants of HyperGCN \mbox{\cite{gcn}} mirroring our two hypergraph views, namely HyperGCN-Demd (demand-based view) and HyperGCN-Dist (distance-based view). Their hyperedge configurations when building hypergraphs follow the same settings used in our HyperCast. All implemented baselines adhere to the original papers' architectures and default recommended hyperparameters. Note that our HyperCast does not require any additional data compared to existing GNN-based approaches for spatiotemporal forecasting.

We use MSE, mean absolute error (MAE), and the coefficient of determination (also referred to as R$^2$) as metrics to evaluate the forecast performance. For the former two, lower error values indicate more accurate predictions while higher values of R$^2$ reflect better forecasts.

\begin{figure*}[!t]
    \centering
    \includegraphics[width=.85\linewidth]{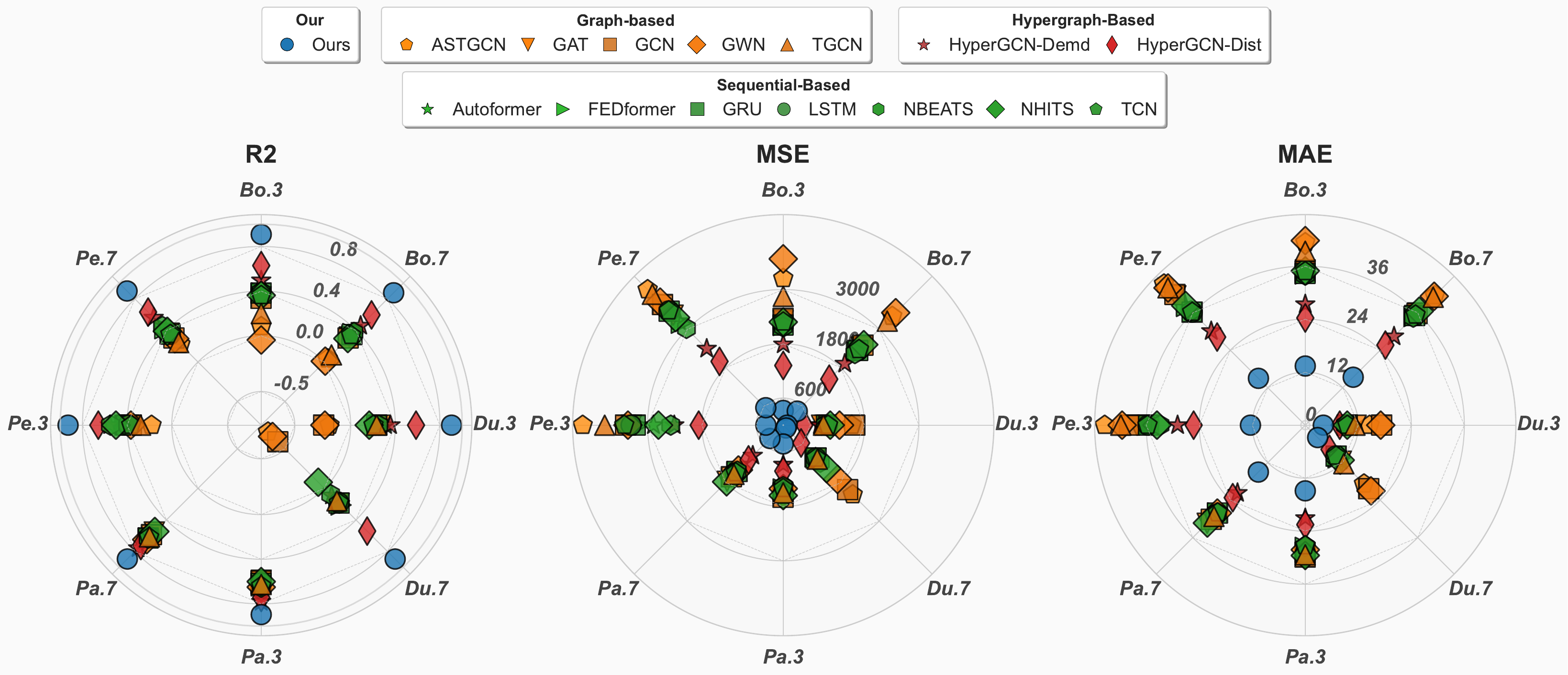}
    \caption{Comparative illustration of forecast performance across our model and all baselines. }
    \label{fig:performance_radar}
\end{figure*}

\subsection{Performance Comparison Against Baselines}
\label{sec:baseline_comparison}

To assess the efficacy of our HyperCast model, we conduct a comprehensive comparison against a suite of aforementioned baseline methods. Our evaluation spans four used datasets and two forecast horizons: $T_f=3$ and $T_f=7$.

First, we present the quantitative results of the Palo Alto dataset in Table \ref{tab:benchmark_comp} as an example. For the three-day-ahead forecast horizon, our HyperCast model (denoted as ``Ours'' in the table) demonstrates its effectiveness by achieving an MSE of $830$, an MAE of $21.3$ kW, and an $\mathrm{R}^2$ of $0.80$ even in its worst-performing configuration, with the best-performing configuration reaching an MSE of $416$, an MAE of $14.8$ kW, and an $\mathrm{R}^2$ of $0.89$. The performance variation is attributed to different hyperparameter settings, i.e., \textit{the number of hyperedges} and \textit{look-back horizons}. These results, derived from all tested hyperparameter combinations, are substantially lower than those of all baseline models, indicating significantly better forecast accuracy. As presented in Table \ref{tab:benchmark_comp}, the superior performance of HyperCast on the Palo Alto dataset remains consistent for the longer forecast horizon of $T_f=7$. 

Interestingly, several sequential models with more advanced architectures, such as DeepAR, Informer, and DLinear, produce unsatisfactory results, underperforming even simpler models like LSTM and GRU. In contrast, while not matching HyperCast, other graph-based methods consistently deliver more reasonable outcomes. This observation may suggest an inherent advantage of graph-based neural structures in capturing the complex spatial patterns of EV charging demand, especially within an urban region. Notably, both HyperGCN variants outperform the graph-based (e.g., GCN and GAT) and sequence-based (e.g., LSTM and Autoformer) baselines, indicating a clear advantage from introducing hypergraph structure over pairwise graphs or purely temporal models.

A broader visual comparison across all four datasets (abbreviated as \textit{Bo.}~for Boulder, \textit{Du.}~for Dundee, \textit{Pa.}~for Palo Alto, and \textit{Pe.}~for Perth) and both forecast horizons is provided in Fig. \ref{fig:performance_radar}. For example, three-day-ahead forecasting at Boulder is written as \textit{Bo.3}. In this radar plot, smaller values closer to the center circles indicate better forecast accuracy for MSE and MAE, whereas larger values further from the center indicate better forecast accuracy for R2. In Fig. \ref{fig:performance_radar}, we select the HyperCast model configuration that achieves the best performance, which is depicted as blue circle in Fig. \ref{fig:performance_radar}.  These comprehensive results further affirm the substantial improvements offered by our HyperCast, which consistently outperforms a wide array of forecasting techniques across diverse datasets and forecast horizons. Further, the results demonstrate that HyperCast is not tailored to a single city but generalizes effectively across different charging infrastructure layouts. To ensure reproducibility of the results in Fig. \mbox{\ref{fig:performance_radar}}, we, in Table \mbox{\ref{tab:optimal_config}}, report the optimal hyperparameter configurations of HyperCast for each dataset under both three-day-ahead ($T_f=3$) and one-week-ahead ($T_f=7$) forecasting scenarios. The table specifies the number of hyperedges ($K$), the recent look-back horizon ($T_r$), and the weekly look-back horizon ($T_w$) that lead to the best performance in each case. A more detailed analysis of hyperparameter effects on forecasting performance is provided in the following subsection.

\begin{figure}[!t]
    \centering
    \includegraphics[width=\linewidth]{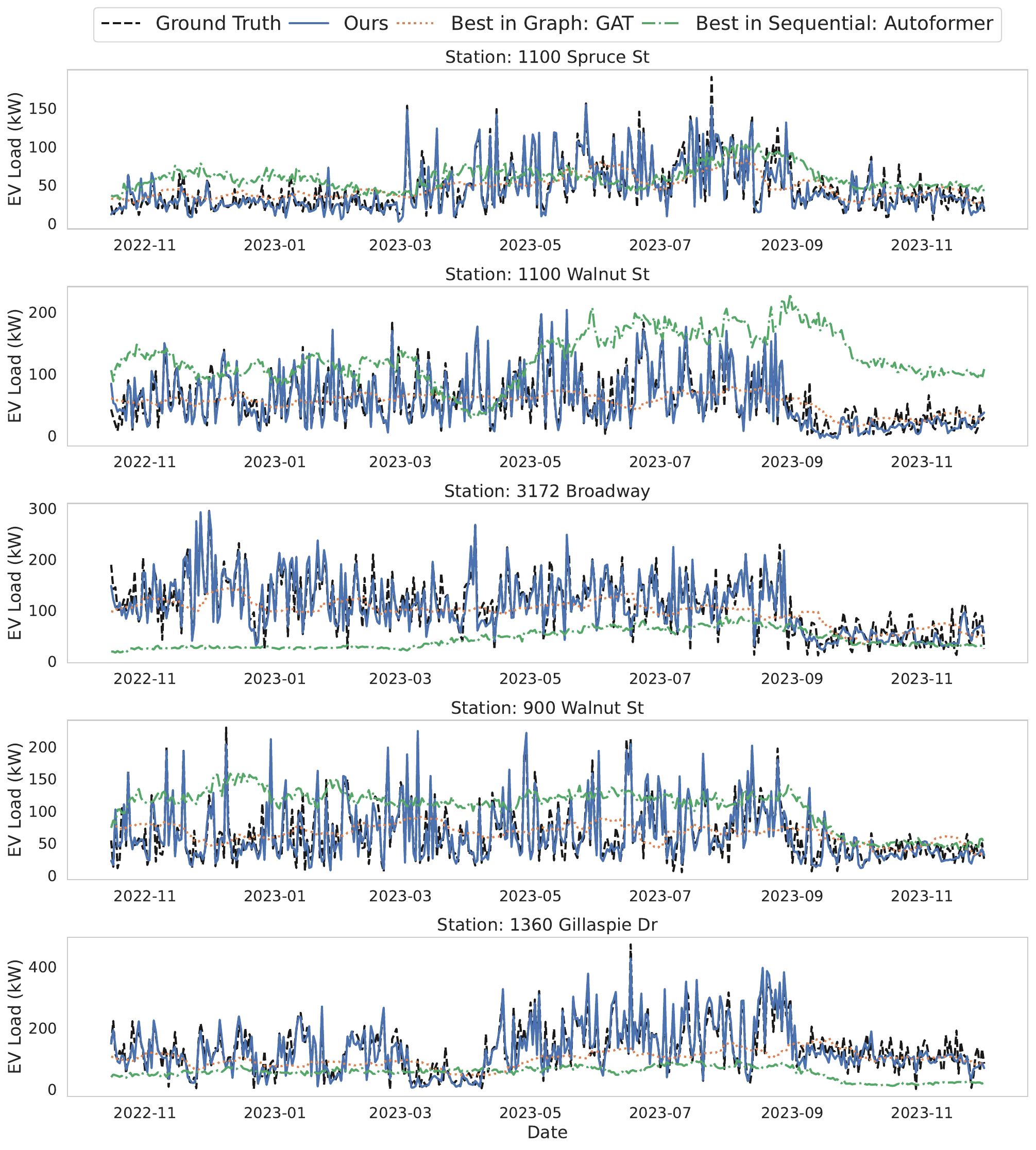}
    \caption{Time-series visualization of predictions and ground-truth demands.}
    \label{fig:prediction_visualization_example}
\end{figure}

We further provide the ground truth demand curves for the Boulder dataset in Fig. \mbox{\ref{fig:prediction_visualization_example}}, alongside predictions from HyperCast, the best-performing graph-based baseline (i.e., GAT), and the best-performing sequential-based baseline (i.e., AutoFormer). From Fig. \mbox{\ref{fig:prediction_visualization_example}}, the EV charging demand curves are highly irregular, with sharp peaks, sudden drops, and significant heterogeneity across stations. Our HyperCast tracks both the magnitude and timing of volatile peaks, such as at Station 3172 Broadway and Station 1360 Gillaspie Dr, more accurately than the baselines. The results visually highlight the advantage of HyperCast in capturing both fine-grained fluctuations and broader demand trends.

Regarding the model computation costs, despite the complex neural architecture design (compared to standard GNN-based methods), the inference times per time step are all under $50$ milliseconds across four datasets, suggesting that the HyperCast's complexity does not translate into significant real-world slowdown for the daily EV charging demand forecasting task.

\begin{table}[!t]
\centering
\caption{Optimal hyperparameter configurations for each dataset under three-day-ahead and one-week-ahead forecasting scenarios.}
\begin{tabular}{c ccc || ccc  }

\toprule

\multirow{2}{*}{}  & \multicolumn{3}{c}{$T_f=3$} & \multicolumn{3}{c}{$T_f=7$}  \\

\cmidrule(lr){2-4} \cmidrule(lr){5-7}

 & \multicolumn{1}{c}{$K$} & \multicolumn{1}{c}{$T_r$} & \multicolumn{1}{c}{$T_w$} & \multicolumn{1}{c}{$K$} & \multicolumn{1}{c}{$T_r$} & \multicolumn{1}{c}{$T_w$} \\

\midrule

Boulder  & $5$ & $7$ & $3$ & $4$ & $21$ & $1$  \\

Dundee  & $25$ & $14$ & $4$ & $20$ & $28$ & $2$  \\

Palo Alto & $5$ & $28$ & $3$ & $5$ & $14$ & $2$  \\

Perth  & $9$ & $14$ & $3$ & $9$ & $28$ & $3$ \\

\bottomrule
\end{tabular}

\label{tab:optimal_config}
\end{table}

\begin{figure}[!t]
    \centering
    \subfloat[MSE]{
    \includegraphics[width=.45\linewidth]{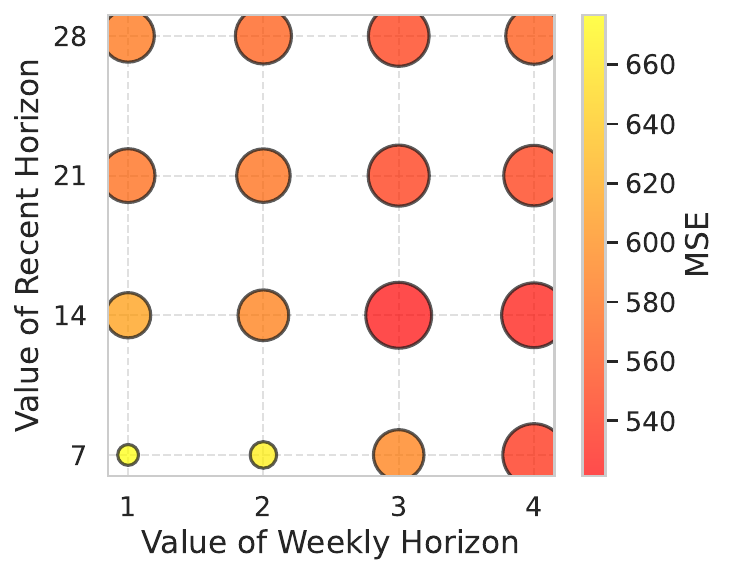}
    \label{fig:param_impact_recent_weekly_mse}
    }
    \subfloat[$\mathrm{R}^2$]{
    \includegraphics[width=.45\linewidth]{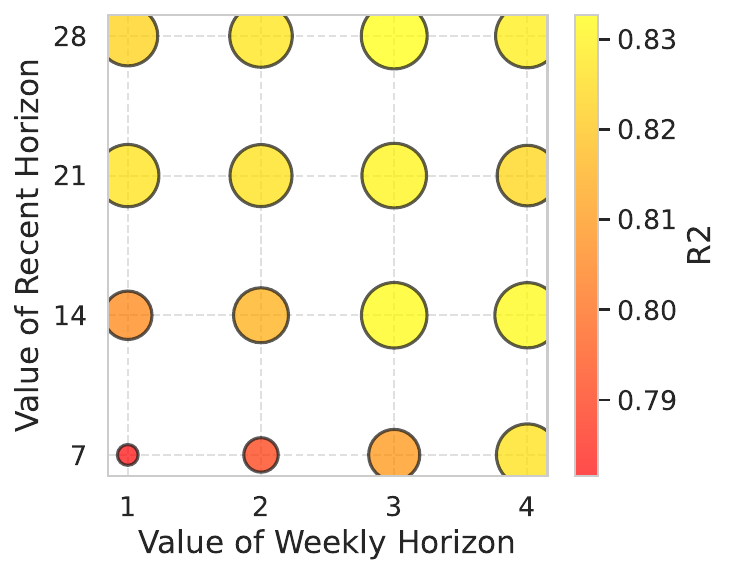}
    \label{fig:param_impact_recent_weekly_r2}
    }
    \caption{Hyperparameter impact of recent and weekly horizons.}
    \label{fig:param_impact_recent_weekly}
\end{figure}

\subsection{Hyperparameter Impact on Forecast Performance}

In this subsection, we aim to analyze the impacts of key hyperparameters on our HyperCast's forecast performance, including the number of hyperedges ($K$), the look-back horizons for the recent time-scale ($T_r$) and the weekly time-scale ($T_w$), which to a great extent determines the quality of learned spatiotemporal charging patterns.

Fig. \ref{fig:param_impact_recent_weekly} illustrates the impacts of varying look-back horizons on forecast performance, where Fig. \ref{fig:param_impact_recent_weekly_mse} and \ref{fig:param_impact_recent_weekly_r2} present corresponding MSE and R$^2$ results. Note that these metrics presented in these figures for each combination of $T_r$ and $T_w$ are an average across the different configurations of $K$ used in the experiments. From Fig. \ref{fig:param_impact_recent_weekly_mse}, lower MSE values (approximately ranging from $540$ to $560$) are generally achieved when the recent horizon is set to $14$ or $21$ days. The weekly horizon $T_w$ exhibits a less significant individual impact; however, very short weekly horizons (e.g., one week) combined with either very long or short recent horizons tend to yield slightly higher MSE. The R$^2$ results shown in Fig. \mbox{\ref{fig:param_impact_recent_weekly_r2}} further align with and reinforce the conclusions drawn from Fig. \mbox{\ref{fig:param_impact_recent_weekly_mse}}. Based on the aggregated results across varying $K$ and all four datasets, the combination of $T_r=14$ and $T_w=3$ emerges as a particularly effective configuration, consistently yielding one of the lowest MSEs and highest R$^2$ values.

\begin{figure*}[!t]
    \centering
    \subfloat[$K$ vs. $T_r$ -- Boulder]{
    \includegraphics[width=.22\linewidth]{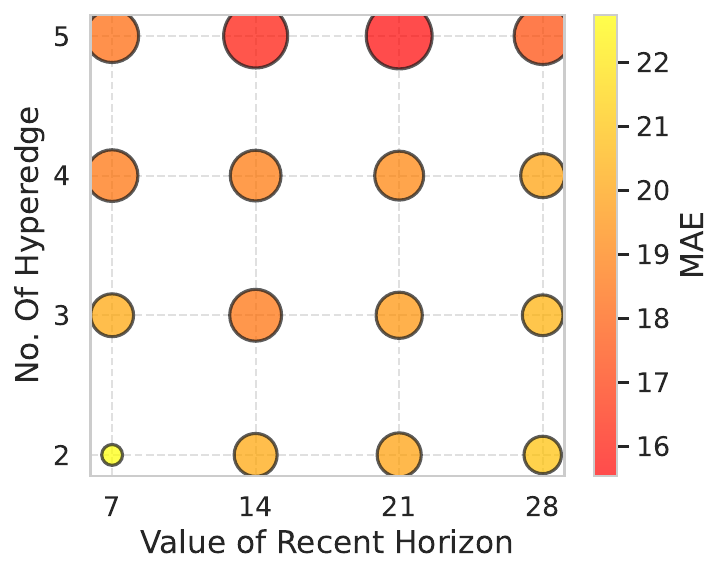}
    \label{fig:param_impact_numhyperedge_recent_boulder}
    }
    \subfloat[$K$ vs. $T_r$ -- Dundee]{
    \includegraphics[width=.22\linewidth]{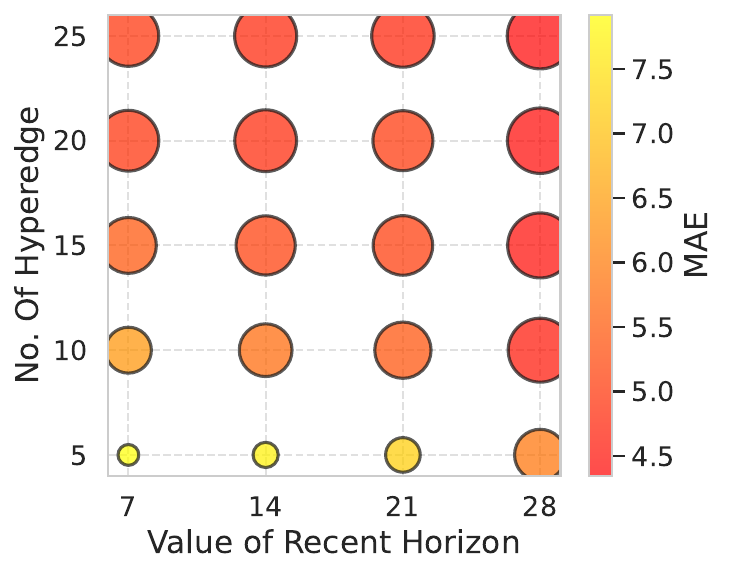}
    \label{fig:param_impact_numhyperedge_recent_dundee}
    }
    \subfloat[$K$ vs. $T_r$ -- Palo Alto]{
    \includegraphics[width=.22\linewidth]{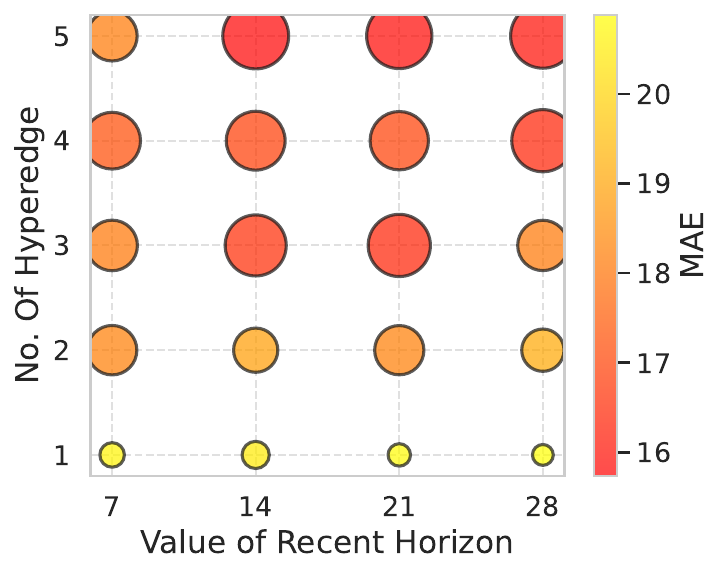}
    \label{fig:param_impact_numhyperedge_recent_paloalto}
    }
    \subfloat[$K$ vs. $T_r$ -- Perth]{
    \includegraphics[width=.22\linewidth]{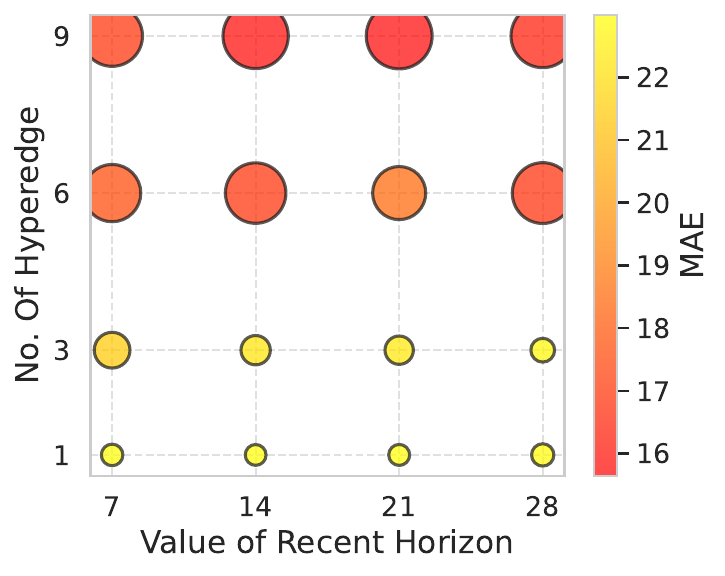}
    \label{fig:param_impact_numhyperedge_recent_perth}
    }\\
    \subfloat[$K$ vs. $T_w$ -- Boulder]{
    \includegraphics[width=.22\linewidth]{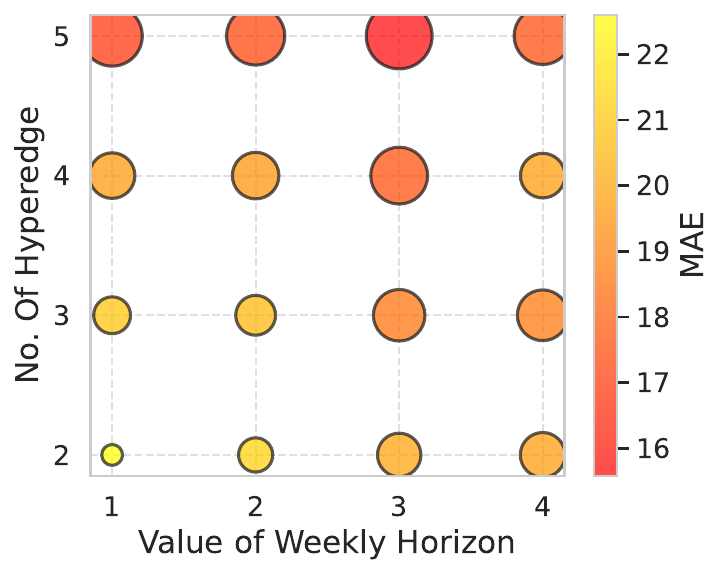}
    \label{fig:param_impact_numhyperedge_weekly_boulder}
    }
    \subfloat[$K$ vs. $T_w$ -- Dundee]{
    \includegraphics[width=.22\linewidth]{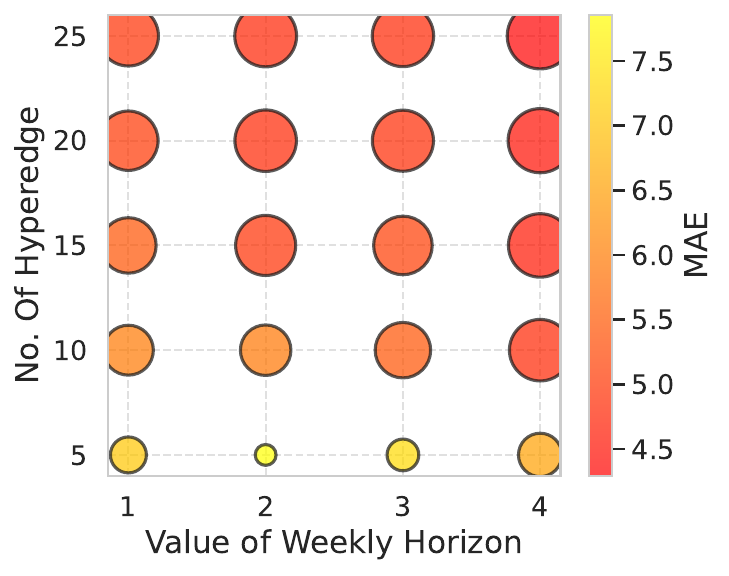}
    \label{fig:param_impact_numhyperedge_weekly_dundee}
    }
    \subfloat[$K$ vs. $T_w$ -- Palo Alto]{
    \includegraphics[width=.22\linewidth]{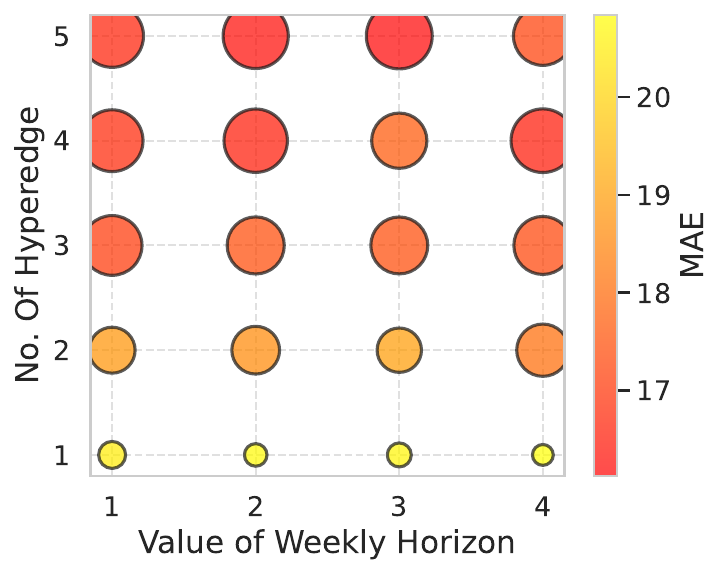}
    \label{fig:param_impact_numhyperedge_weekly_paloalto}
    }
    \subfloat[$K$ vs. $T_w$ -- Perth]{
    \includegraphics[width=.22\linewidth]{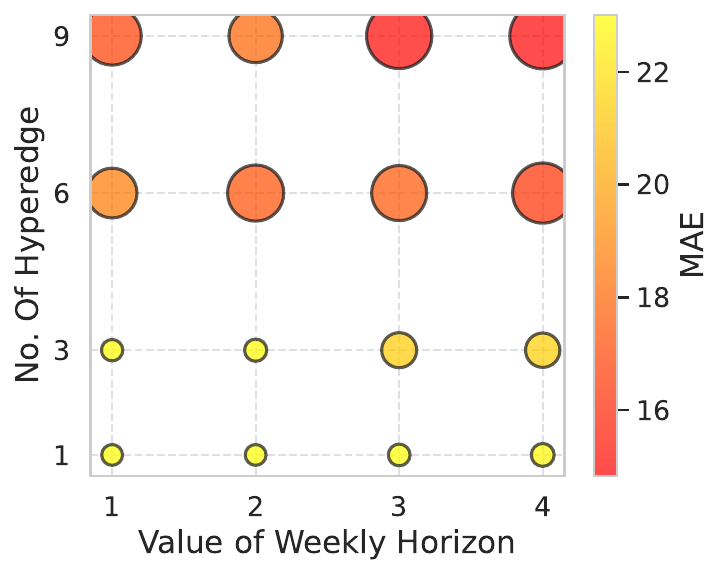}
    \label{fig:param_impact_numhyperedge_weekly_perth}
    }\\
    \caption{Hyperparameter impacts of look-back horizons and the number of hyperedges.}
    \label{fig:param_impact_numhyperedge_recent_weekly}
\end{figure*}

Furthermore, Fig. \ref{fig:param_impact_numhyperedge_recent_weekly} depicts the effects of varying the number of hyperedges in conjunction with both recent and weekly look-back horizons, using MAE as the evaluation metric. Surprisingly, for look-back horizons, the general trend derived from the averaged results across datasets and metrics is not uniformly ``more is better''; instead, specific ranges show better performance. For the recent look-back horizon, a moderately long history appears most beneficial, while for the weekly look-back horizon, extending the history beyond a single instance is favored. More specifically, a recent look-back horizon of $14$ or $21$ days frequently appears optimal among all configurations. This range appears to strike a balance between capturing sufficient historical variation to learn medium-term demand shifts, while avoiding the redundancy and noise that longer horizons (e.g., 28 days) may introduce. Moreover, a two- to three-week window aligns with common behavioral and activity cycles, such as commuting routines and biweekly events, which naturally shape EV charging demand. For the weekly look-back horizon $T_w$, values greater than one (i.e., $2$, $3$, or $4$) generally lead to improved performance. Besides, Fig. \mbox{\ref{fig:param_impact_numhyperedge_recent_weekly}} reveals that the increase of hyperedge number has a clearly positive impact on more accurate forecast performance, suggesting that a more granular spatial representation improves forecasting. Therefore, the optimal strategy for setting $K$ is to use the largest reasonable value for each specific dataset's network size: $K=5$ for Boulder, $K=25$ for Dundee, $K=5$ for Palo Alto, and $K=9$ for Perth.

\subsection{HyperCast Ablation Study}

To understand the contribution of different components of the HyperCast model, we conduct an ablation study evaluating several model variants: 1) \textit{wo-HSTB}: HSTB replaced with a graph convolutional block; 2) \textit{wo-Rec.}: recent timescale input removed; 3) \textit{wo-Wek.}: weekly timescale input removed; 4) \textit{wo-Dist.}: distance-based hypergraph view removed; and 5) \textit{wo-Demd.}: demand-based hypergraph view removed.

\begin{figure}[!t]
    \centering
    \includegraphics[width=.85\linewidth]{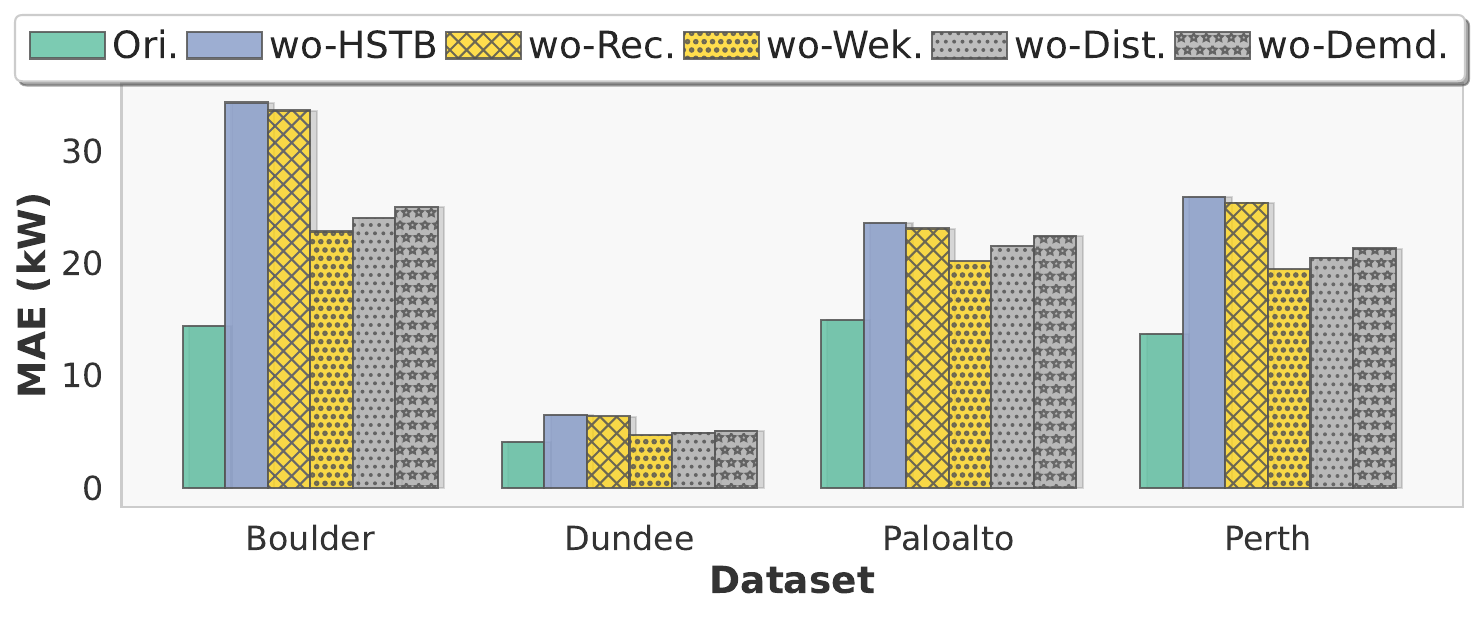}
    \caption{MAE results of the ablation study.}
    \label{fig:ablation_res}
\end{figure}

The MAE for these ablated models and the original HyperCast (referred to as ``\textit{Ori.}'') is presented in Fig.~\ref{fig:ablation_res}. It is noteworthy that the \textit{wo-HSTB} variant shows the largest performance degradation across all datasets, highlighting the crucial role of HSTB in spatiotemporal processing.

Moreover, removing either timescales significantly increases MAE. In particular, the weekly timescale generally leads to less significant impact than the recent timescale, suggesting the importance of short-term dynamics. Similarly, the absence of either hypergraph views causes noticeable MAE increase, indicating the values of considering both geographical proximity and functional similarities from demand patterns, while in datasets like Boulder and Perth, removing the demand-based view has a slightly greater negative impact on forecast accuracy.

Further, to analyze the effect of hypergraph soft assignments, we replace the soft station–hyperedge memberships with rigid one-hot assignments. As reported in Table \mbox{\ref{tab:soft_vs_rigid_ablation}}, rigid memberships consistently degrade performance, with MAE increasing by about $30\%$ relative to the soft assignment. Note that the degradation is most significant on denser, functionally heterogeneous charging networks, e.g., Palo Alto and Dundee, exhibit increases of approximately $33\%$-$35\%$, where stations naturally participate in overlapping groups; even in more dispersed settings such as Boulder and Perth, MAE still rises by about $27\%$ - $29\%$. These results support our design choice: soft memberships enable stations to share multiple hyperedges, preserving cross-group message passing and enabling HS-GAT to reweight group participation contextually by view and timescale, which is crucial for accurate spatiotemporal forecasting. Also, to validate effectiveness of our feature fusion module (i.e., CVF and CTF), we replace these two components with two simpler fusion techniques, including addition and concatenation. The results reveal significant increase (more than $80\%$) of MAE, indicating substantial degradation of forecasting performance.

\begin{table}[!t]
    \centering
    \caption{MAE increase (in percentage) when using rigid hyperedge membership assignment.}
    \resizebox{\columnwidth}{!}{
    \begin{tabular}{c|c|c|c|c}
    
    \hline
    \diagbox[height=2\normalbaselineskip]{$T_f$}{Dataset}  & Boulder & Dundee & Palo Alto & Perth \\

    \hline 

    $T_f=3$ & $27.4\% \boldsymbol{\uparrow}$ & $34.6\% \boldsymbol{\uparrow}$ & $34.5\% \boldsymbol{\uparrow}$ & $29.0\% \boldsymbol{\uparrow}$ \\

    \hline  

    $T_f=7$ & $26.8\% \boldsymbol{\uparrow}$ & $31.4\% \boldsymbol{\uparrow}$ & $32.6\% \boldsymbol{\uparrow}$ & $28.6\% \boldsymbol{\uparrow}$ \\ 
    
    \hline 
    
    \end{tabular}
    }
    
    \label{tab:soft_vs_rigid_ablation}
\end{table}

\subsection{HyperCast Inner Mechanism Analysis}

To analyze why the HyperCast on the EV charging load forecasting task, we here aim to explore the model's inner mechanism to see how it effectively extracts and leverages spatiotemporal EV charging patterns.

\subsubsection{Hyperedge Attention} 
\begin{figure}[!t]
    \centering
    \includegraphics[width=.85\linewidth]{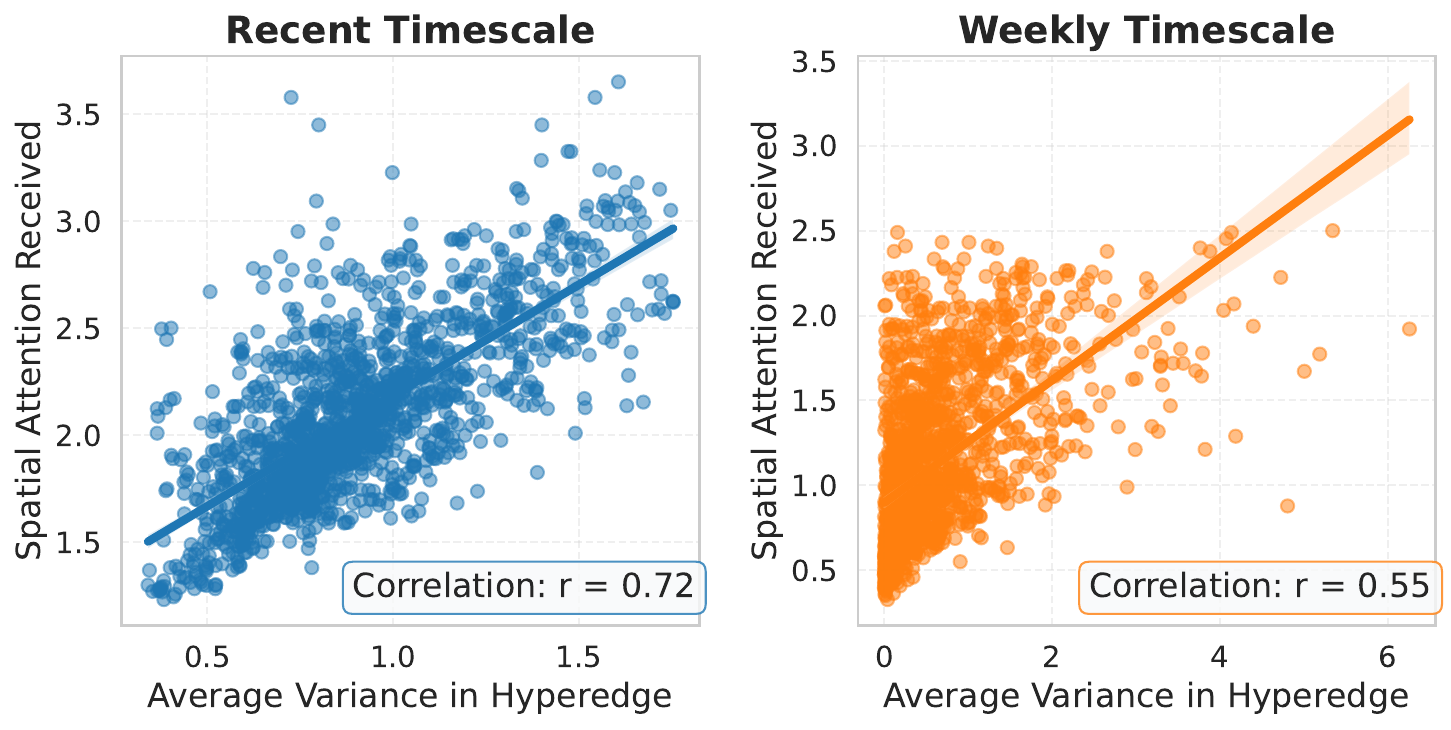}
    \caption{Hyperedge variance vs. corresponding received spatial attention.}
    \label{fig:hyperedge_attn}
\end{figure}

As outlined in Section \ref{subsec:method_HSTB}, the HyperCast first processes spatial information via the HS-GAT. In Fig. \ref{fig:hyperedge_attn}, we observe that if charging demand on one hyperedge fluctuates more (i.e., showing higher variance), the spatial attention received by that hyperedge, expressed as $\sum_{k'\in\mathcal{N}_k}\alpha_{bt'k'k}^{(h)}$ for the $k$-th hyperedge in Eq. \eqref{eq:gat_attn_weight}, increases with a significant positive correlation particularly for the recent timescale (with a correlation coefficient of $0.72$). The average variance of a hyperedge is an aggregated result of each charging station's load variance along with their membership on the specific hyperedge using the calculated hyperedge incidence matrix. The results in Fig. \ref{fig:hyperedge_attn} suggests that the model adaptively focuses on hyperedges with more dynamic and potentially informative patterns.

\begin{figure}[!t]
    \centering
    \subfloat[Hyperedge attention matrix.]{
    \includegraphics[width=.43\linewidth]{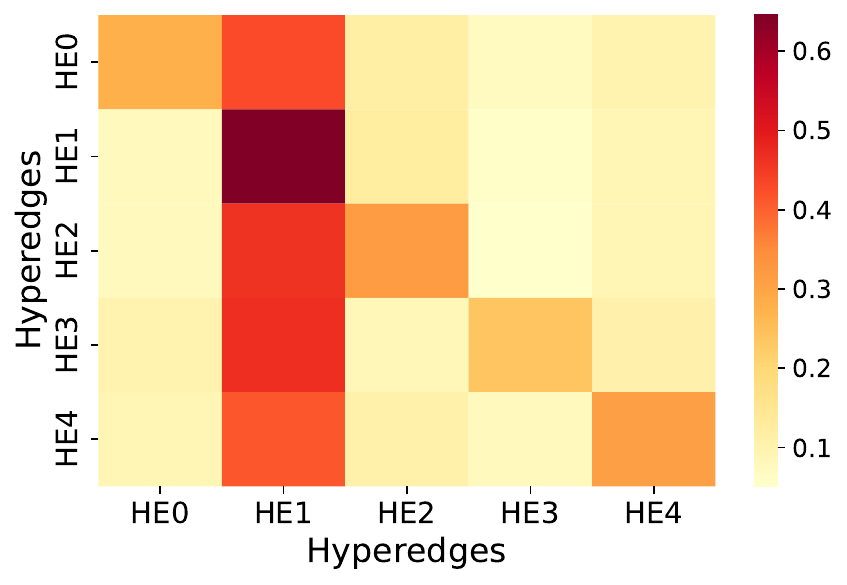}
    \label{fig:example_hyperedge_attn_attn_mat}
    }
    \subfloat[Hyperedge average variance.]{
    \includegraphics[width=.43\linewidth]{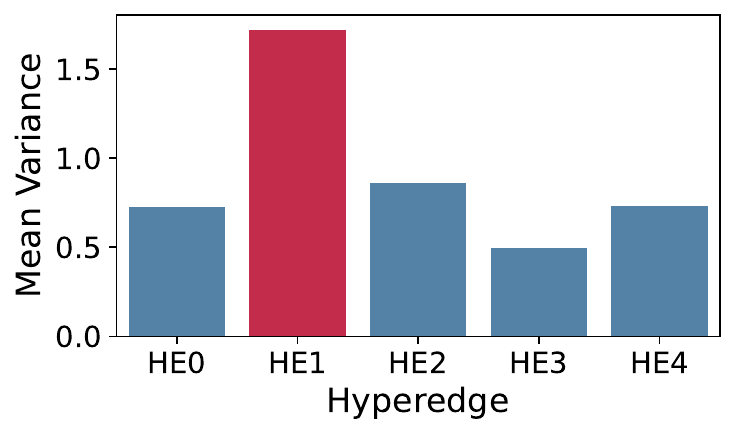}
    \label{fig:example_hyperedge_attn_avg_var}
    }
    \caption{Palo Alto's example of spatial hyperedge attention matrix.}
    \label{fig:example_hyperedge_attn}
\end{figure}

An example from the Palo Alto dataset, shown in Fig. \ref{fig:example_hyperedge_attn}, provides a concrete illustration. Fig. \ref{fig:example_hyperedge_attn_avg_var} displays the mean variance for five hyperedges (denoted as \textit{HE0}-\textit{HE4}), where \textit{HE1} has the highest mean variance compared to others. Fig. \ref{fig:example_hyperedge_attn_attn_mat} presents the corresponding spatial hyperedge attention matrix, where \textit{HE1} receives notable attention from other hyperedges, enabling the model to prioritize regions with greater uncertainty or more significant signals.

More importantly, this attention information could have real-world impacts and can be used by the grid operator to take proactive measures in advance to ensure the grid reliability. For example, the grid operator can allocate more monitoring resources and maintain a higher state of alert for these specific ``hotspots''. Also, dynamic pricing schemes or demand-response programs in these areas can be implemented to incentivize off-peak charging and smooth out volatile demand peaks.

\begin{figure}[!t]
    \centering
    \subfloat[Boulder]{
    \includegraphics[width=.48\linewidth]{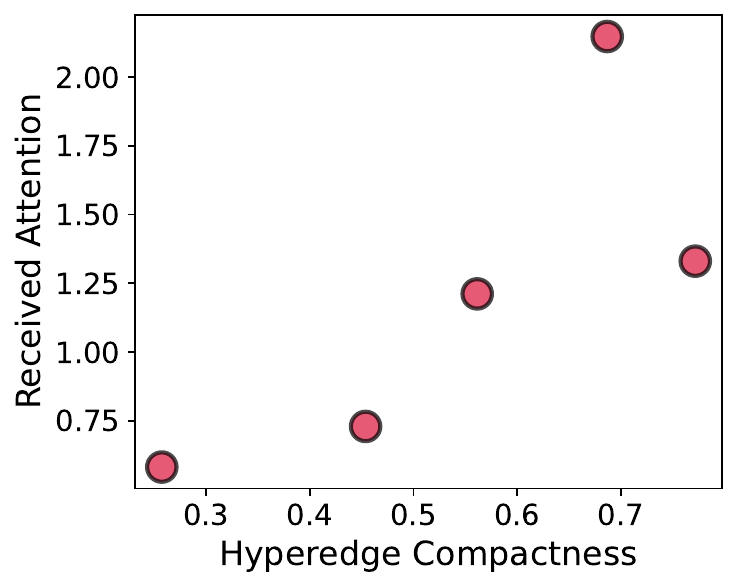}
    \label{fig:hyperedge_compactness_Boulder}
    }
    \subfloat[Dundee]{
    \includegraphics[width=.48\linewidth]{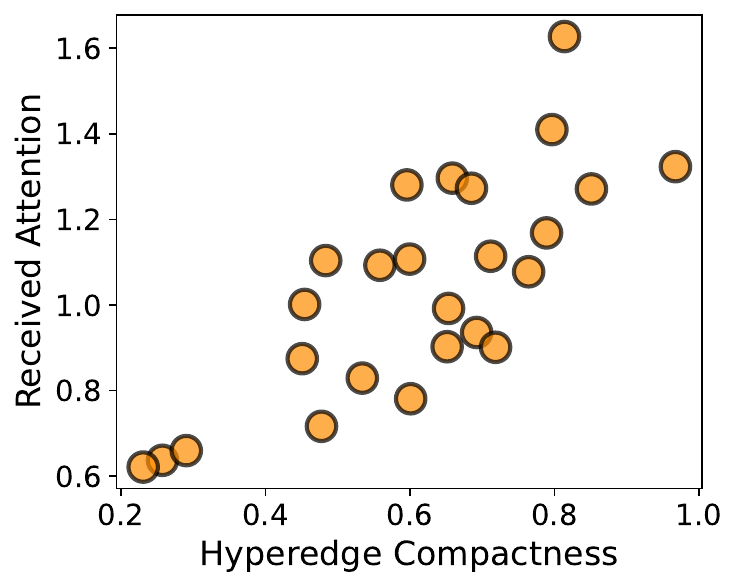}
    \label{fig:hyperedge_compactness_Dundee}
    } \\ 
    \subfloat[Palo Alto]{
    \includegraphics[width=.48\linewidth]{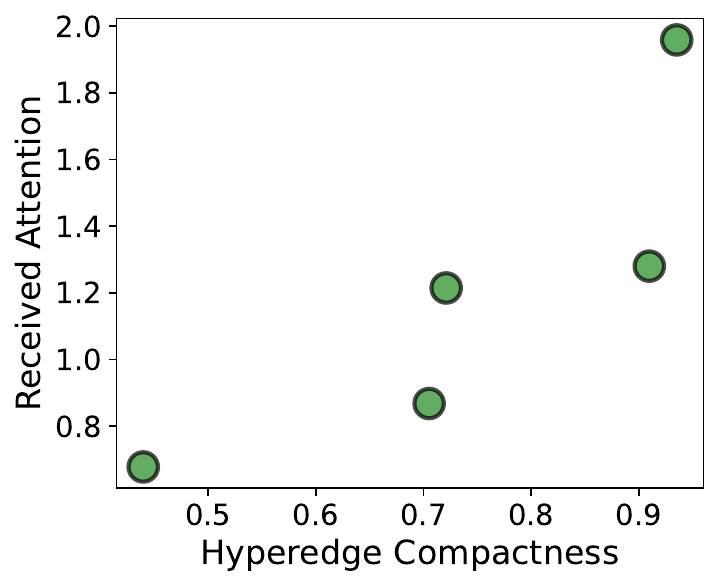}
    \label{fig:hyperedge_compactness_PaloAlto}
    }
    \subfloat[Perth]{
    \includegraphics[width=.48\linewidth]{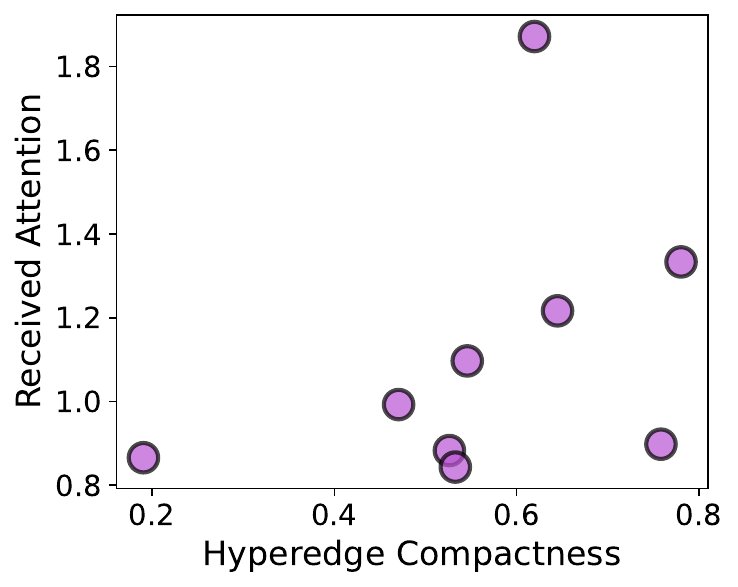}
    \label{fig:hyperedge_compactness_Boulder_Perth}
    }
    \caption{Visualization between hyperedge compactness and received attention.}
    \label{fig:hyperedge_compactness}
\end{figure}

In addition to demand-based view, we further analyze the hyperedge attention from the distance-based view. Specifically, we calculate hyperedge compactness (by averaging, inverting, and normalizing the pairwise distance within each hyperedge) and then illustrate the relationships between hyperedge compactness and hyperedge received attention in Fig. \mbox{\ref{fig:hyperedge_compactness}}. The results indicate that more compact hyperedges receive higher attention. From an intuitive perspective, when a hyperedge forms a well–localized neighborhood, signals from its members are more mutually informative. Our HS-GAT therefore focuses on these hyperedges more when aggregating spatial information.

\subsubsection{Temporal Attention} 
\begin{figure}[!t]
    \centering
    \includegraphics[width=.9\linewidth]{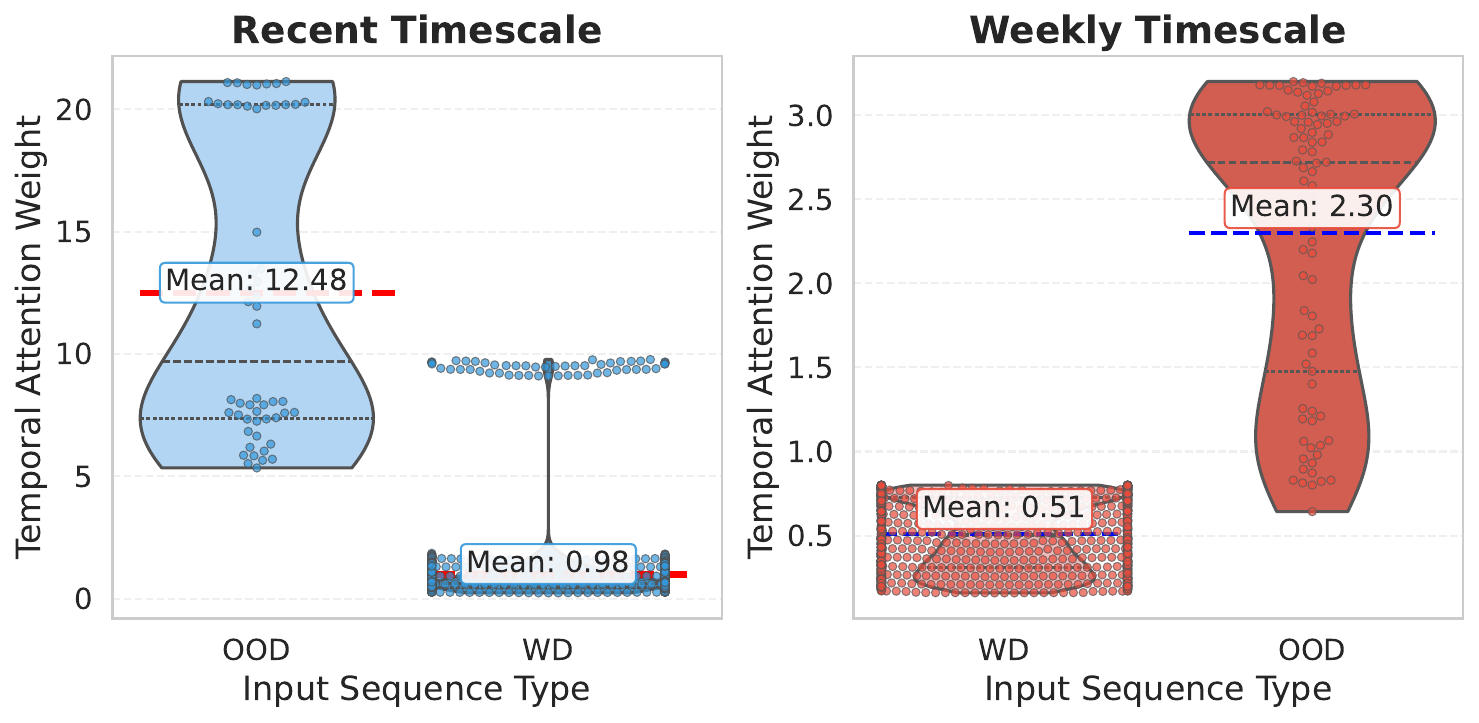}
    \caption{Temporal attention of time steps w.r.t. OOD and WD charging demand.}
    \label{fig:HSTB_temp_attn}
\end{figure}

After the HS-GAT, a TTE extracts temporal patterns from the input time-series EV charging demand. The temporal attention weight is calculated as per Eq. \eqref{eq:HSTB_temp_attn_weight}. For the input time-series from either the recent or the weekly time-scales, we calculate each time-series' mean and standard deviation, then identify those time steps whose charging demand is out-of-distribution (OOD) based on the \textit{three-sigma} rule. Others are referred to as within-distribution (WD) time steps. The results of received temporal attention for OOD and WD time steps are illustrated in Fig. \ref{fig:HSTB_temp_attn}, indicating that OOD time steps receive significantly higher temporal attention than WD time steps across both timescales. Specifically, for the recent timescale, OOD steps have a mean attention weight of $12.48$, starkly contrasting with $0.98$ for WD steps. Similarly, for the weekly timescale, OOD steps attract a mean attention of $2.30$, compared to $0.51$ for WD steps. Such differences indicates that the HyperCast effectively identifies and highlights unusual or anomalous data points within the temporal sequences, which are often more informative for accurate forecasting.

\subsubsection{View and Timescale Attention} 
\begin{figure}[!t]
    \centering
    \includegraphics[width=.75\linewidth]{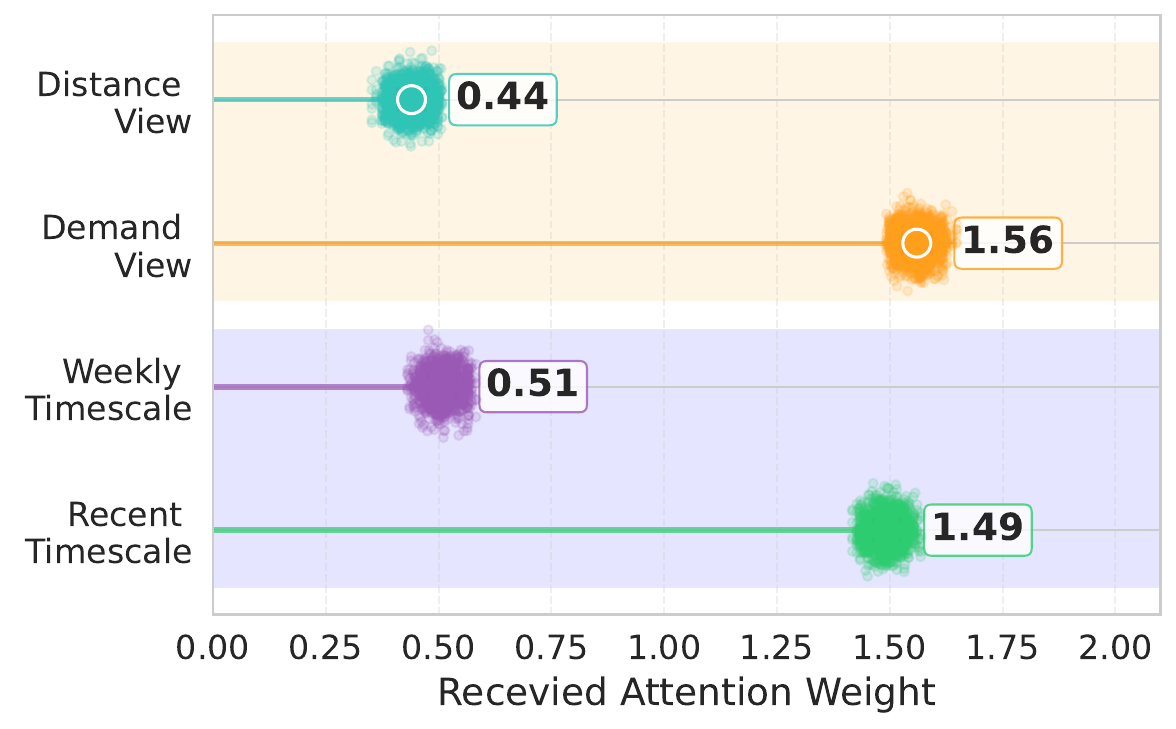}
    \caption{Received time-scale and view attention.}
    \label{fig:CVF_CTF_attn}
\end{figure}

Following several HSTBs, the model employs fusion techniques such as CVF and CTF detailed in Section \ref{subsec:method_CVF} and \ref{subsec:method_CTF}, respectively. Analogous to previous subsections, we calculate the attention received by each time-scale within the CTF and each hypergraph view within the CVF. Note that the sum of received attention across all views or timescales equals two. The results are presented in Fig. \ref{fig:CVF_CTF_attn}. In CTF, the recent timescale receives a substantially higher attention weight ($1.49$) compared to the weekly timescale ($0.44$), suggesting that more recent information is weighted more heavily when fusing temporal contexts. In CVF, the demand-based view attracts a higher attention weight ($1.56$) than the distance-based view ($0.51$), implying that functional similarities captured by the demand-based hypergraphs are given more importance than geographical proximity when fusing view-specific information. These differential attention weights demonstrate how HyperCast adaptively prioritizes different information sources during its fusion stages. In addition to the Palo Alto dataset, Dundee converges to demand~1.53 vs. distance~0.47 and recent~1.46 vs. weekly~0.54, indicating dominance of functional similarity and recent dynamics. Boulder reaches a more balanced split—demand~1.18 vs. distance~0.82 and recent~1.08 vs. weekly~0.92—reflecting mixed variability and a less concentrated geography. Perth shows only a mild preference toward demand and recent (demand~1.08 vs. distance~0.92; recent~1.05 vs. weekly~0.95), consistent with a medium-density network in which spatial proximity and weekly cycles remain comparatively informative. Such trends suggest that, in dense urban charging networks, HyperCast increasingly favors demand-based interactions and short-term dynamics, whereas in more dispersed contexts, geography and weekly regularities retain a larger role.

As shown in Fig. \mbox{\ref{fig:CVF_CTF_attn}}, the demand-based view and recent timescale consistently receive higher attention weights across the urban datasets studied in this work. This outcome is intuitive, as functional similarities in charging behavior often provide stronger predictive signals than geographic proximity in dense networks, and recent demand dynamics typically capture more relevant short-term fluctuations than longer-term periodic patterns. However, this prioritization is not expected to be universal. In more sparsely populated rural charging networks, where stations are geographically isolated and functional similarity is weak, the distance-based view may carry greater importance by capturing catchment-area separations and substitution effects. Likewise, in systems dominated by highly regular weekly patterns, the weekly timescale could outweigh the recent horizon in predictive value. These considerations highlight that HyperCast’s fusion mechanism is adaptive by design: it can dynamically adjust the relative importance of different views and timescales depending on the structural and behavioral characteristics of the underlying charging network.

\subsubsection{Station-Wise Attentions} 
\begin{figure}[!t]
    \centering
    \subfloat[Recent, Demand-based]{
    \includegraphics[width=.49\linewidth]{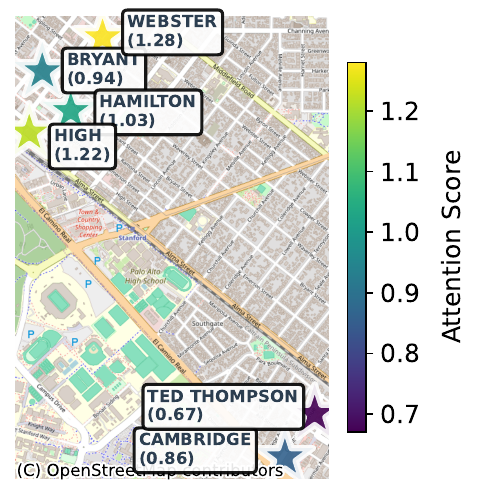} 
    \label{fig:geo_map_with_attn_recDemd}
    } 
    \subfloat[Recent, Distance-based]{
    \includegraphics[width=.49\linewidth]{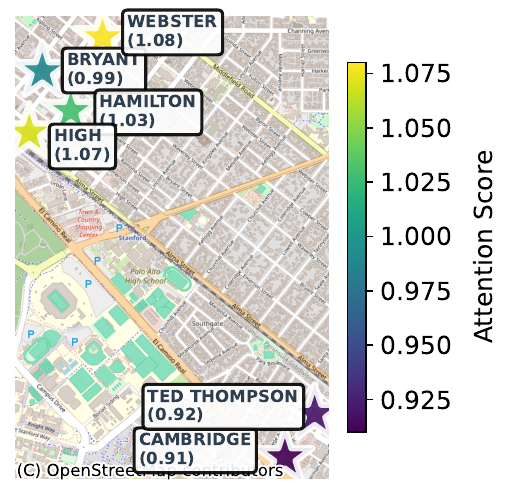}
    \label{fig:geo_map_with_attn_recDist}
    }  \\ 
    \subfloat[Weekly, Demand-based]{
    \includegraphics[width=.49\linewidth]{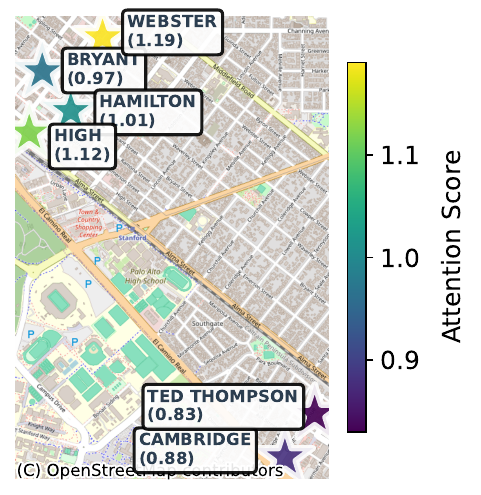}
    \label{fig:geo_map_with_attn_wekDemd}
    } 
    \subfloat[Weekly, Distance-based]{
    \includegraphics[width=.49\linewidth]{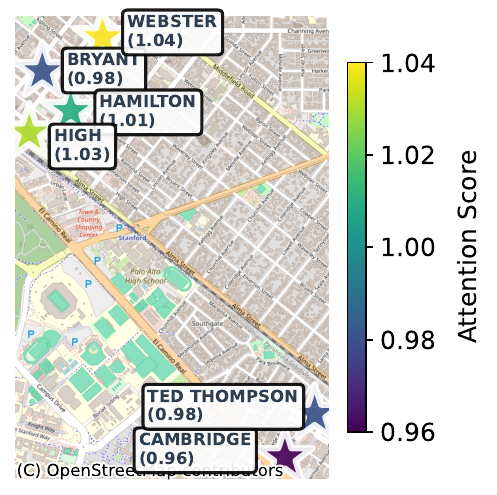}
    \label{fig:geo_map_with_attn_wekDist}
    } 
    \caption{Station-wise attention maps for the Palo Alto dataset.}
    \label{fig:geo_map_with_attn}
\end{figure}

To improve interpretability by linking learned attentions to physical geography, we visualize station–wise received attention on a city basemap. Fig. \mbox{\ref{fig:geo_map_with_attn}} shows four panels for Palo Alto (recent/weekly timescales combined with demand/distance views), where node color encodes how much attention a station receives from all other stations under the corresponding stream.

\begin{figure*}[!t]
    \centering
    \includegraphics[width=\linewidth]{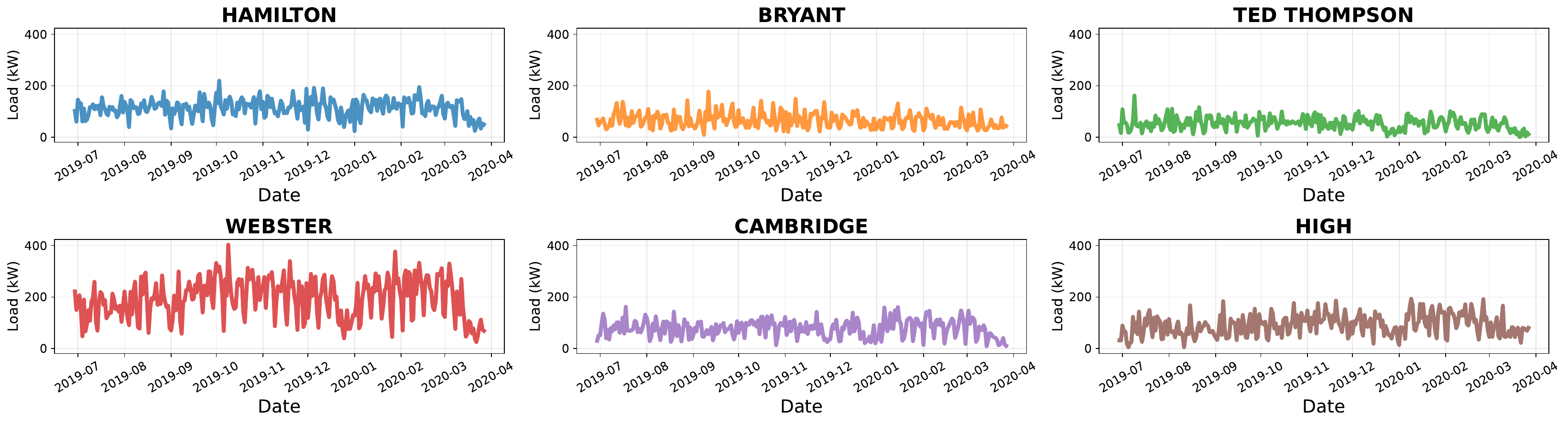}
    \caption{EV charging demand curves of all charging stations in the Palo Alto dataset.}
    \label{fig:charging_demand_curves_PaloAlto}
\end{figure*}

\textit{Recent Timescale \& Demand-based View}: This stream emphasizes short-term volatility and OOD events. Consistent with the load traces (shown in Fig. \mbox{\ref{fig:charging_demand_curves_PaloAlto}}), the downtown business–district stations attract the most attention: WEBSTER and HIGH, followed by HAMILTON and BRYANT, while the south–east pair (i.e., CAMBRIDGE and TED THOMPSON) are lower. The pattern aligns with previous results that hyperedges (and hence stations) exhibiting higher variance receive stronger spatial and temporal attention.

\textit{Recent Timescale \& Distance-based View}: When only spatial proximity is considered, attention concentrates on dense hubs in the downtown cluster. Scores are more uniform than in the demand view -- WEBSTER $1.08$, HIGH $1.07$, HAMILTON $1.03$, BRYANT $0.99$ -- with the south–east pair slightly lower (TED THOMPSON $0.92$ and CAMBRIDGE $0.91$). This reflects connectivity density rather than volatility: closely packed neighbors exchange more attention within the core.

\textit{Weekly Timescale \& Demand-based View}: Weekly modeling smooths transient spikes and emphasizes business-week regularities. The demand-based ranking persists but is compressed: WEBSTER $1.19$, HIGH $1.12$, HAMILTON $1.01$, BRYANT $0.97$, CAMBRIDGE $0.88$, and TED THOMPSON $0.83$. This is consistent with Fig. \mbox{\ref{fig:CVF_CTF_attn}}, where the demand view remains informative while weekly horizons reduce contrast relative to recent horizons.

\textit{Weekly Timescale \& Distance-based View}: This is the most uniform distribution, as geography dominates and weekly aggregation further dampens demand-induced asymmetries: WEBSTER $1.04$, HIGH $1.03$, HAMILTON $1.01$, BRYANT $0.98$, TED THOMPSON $0.98$, and CAMBRIDGE $0.96$. The downtown four retain a small edge as the principal spatial hub, while the south–east pair sit slightly lower due to fewer immediate neighbors.

In summary, the maps corroborate our attention mechanism analysis from the following three perspectives: 1) demand–based views highlight stations with dynamic informativeness (higher variance and OOD events), with the recent horizon accentuating these effects; 2) distance–based views reflect spatial hubness driven by geographic density; and 3) weekly horizons yield tighter spreads by prioritizing periodic structure. These results illustrate how HyperCast’s fusion is context-adaptive: the model concentrates attention on stations that are most informative for the active view/timescale, while still respecting the underlying geography.

\section{Conclusion and Future Work} \label{sec:conclusion}

In this paper, we address the pressing challenge of accurately forecasting EV charging demand by tackling the limitations of existing spatiotemporal models regarding insufficient modeling on group-wise interactions that characterize real-world charging station networks. We develop HyperCast, which is built upon the expressive power of hypergraphs. Our approach features a unique multi-view hypergraph construction methodology, combining static distance-based and dynamic demand-based views with soft assignments to flexibly model higher-order relationships. This spatial representation, extracted by tailored HSTBs, is integrated with multi-timescale inputs to disentangle short-term dynamics from weekly periodicities. Validated on four public datasets, HyperCast confirms its effectiveness in more accurate EV charging demand forecasting.

Our analysis of HyperCast's internal attention mechanisms provides valuable insights into its prediction reasoning process. For example, the model learns to adaptively prioritize information by focusing its spatial attention on hyperedges with higher demand variance, its temporal attention on anomalous or OOD events, and its fusion attention on more informative sources, such as recent timescale data and demand-based functional similarities. These findings explain how HyperCast effectively navigates complex spatiotemporal patterns to achieve more accurate forecasting.

For future work, several promising directions exist. First, incorporating a richer set of exogenous variables, such as real-time traffic, weather conditions, and detailed point-of-interest data, could further enhance forecast performance by providing direct causal context that HyperCast cannot infer from historical charging data alone. Secondly, integrating the HyperCast into downstream optimization and control applications, such as smart charging coordination or vehicle-to-grid management, presents a compelling avenue for future research.

\bibliographystyle{ieeetr}
\bibliography{IEEEabrv}

\end{document}